\documentclass[lettersize, journal]{IEEEtran}
\usepackage[utf8]{inputenc}
\usepackage{hyperref}
\usepackage{graphicx} 
\usepackage{caption} 
\usepackage{multirow}
\usepackage{booktabs}
\usepackage{subcaption}
\usepackage{color}
\usepackage{amsmath}
\usepackage{bm}
\usepackage{amsfonts}
\usepackage{makecell}
\usepackage{xcolor}
\usepackage{newpxtext}
\usepackage{array} 
\usepackage{cite}
\usepackage{bbding}

\renewcommand\arraystretch{1.2} 

\newcommand{\eg}{\emph{e.g.}}
\newcommand{\etal}{\emph{et al.}}
\newcommand{\ie}{\emph{i.e.}}

\UseRawInputEncoding
\begin{document}

\title{GroupedMixer: An Entropy Model with \\ Group-wise Token-Mixers \\ for Learned Image Compression}

\author{Daxin Li, Yuanchao Bai,~\IEEEmembership{Member,~IEEE,} Kai Wang, Junjun Jiang,~\IEEEmembership{Senior Member,~IEEE,}\\  Xianming Liu,~\IEEEmembership{Member,~IEEE,}  Wen Gao,~\IEEEmembership{Fellow,~IEEE} 

\IEEEcompsocitemizethanks{This work was supported in part by National Key Research and Development Program of China under Grant 2022YFF1202104, in part by National Natural Science Foundation of China under Grants 62301188, 92270116, 62071155 and U23B2009, in part by the Strategic Research, and Consulting Project by the Chinese Academy of Engineering under Grant 2023-XY-39, in part by China Postdoctoral Science Foundation under Grant 2022M710958, and in part by Heilongjiang Postdoctoral Science Foundation under Grant LBH-Z22156.}
\IEEEcompsocitemizethanks{
\IEEEcompsocthanksitem Daxin Li, Yuanchao Bai,  Kai Wang, Junjun Jiang and Xianming Liu are with the School of Computer Science and Technology, Harbin Institute of Technology, Harbin, 150001, China, E-mail: \{hahalidaxin, cswangkai\}@stu.hit.edu.cn, \{yuanchao.bai, csxm, jiangjunjun\}@hit.edu.cn.
\IEEEcompsocthanksitem Wen Gao is with the School of Electronics Engineering and Computer Science, Peking University, Beijing, 100871, China, E-mail: wgao@pku.edu.cn. \protect \\
}
\thanks{(Corresponding author: Xianming Liu)}
 }

\markboth{IEEE TRANSACTIONS ON CIRCUITS AND SYSTEMS FOR VIDEO TECHNOLOGY}%
{IEEE TRANSACTIONS ON CIRCUITS AND SYSTEMS FOR VIDEO TECHNOLOGY}

\IEEEpubid{\begin{minipage}{\textwidth}\ \\[6pt] Copyright~\copyright~2024 IEEE. Personal use of this material is permitted. However, permission to use this material for any other purposes must be obtained from the IEEE by sending an email to pubs-permissions@ieee.org\end{minipage}}
\maketitle

\begin{abstract}
Transformer-based entropy models have gained prominence in recent years due to their superior ability to capture long-range dependencies in probability distribution estimation compared to convolution-based methods.
However, previous transformer-based entropy models suffer from a sluggish coding process due to pixel-wise autoregression or duplicated computation during inference.
In this paper, we propose a novel transformer-based entropy model called GroupedMixer, which enjoys both faster coding speed and better compression performance than previous transformer-based methods.
Specifically, our approach builds upon group-wise autoregression by first partitioning the latent variables into groups along spatial-channel dimensions, and then entropy coding the groups with the proposed transformer-based entropy model. The global causal self-attention is decomposed into more efficient group-wise interactions, implemented using inner-group and cross-group token-mixers.
The inner-group token-mixer incorporates contextual elements within a group while the cross-group token-mixer interacts with previously decoded groups. Alternate arrangement of two token-mixers enables global contextual reference. 
To further expedite the network inference, we introduce context cache optimization to GroupedMixer, which caches attention activation values in cross-group token-mixers and avoids complex and duplicated computation.
Experimental results demonstrate that the proposed GroupedMixer yields the state-of-the-art rate-distortion performance with fast compression speed. 
\end{abstract}


\section{Introduction}
\IEEEPARstart{L}{ossy} image compression is a long-standing and vital research topic in signal processing and computer vision fields, which has been widely used to reduce the cost of storage and transmission. Due to the rapidly increasing user demand for online sharing and communication via prevalent applications such as WeChat, Twitter, etc., the number of images on the Internet has grown exponentially. Considering the enormous amount of image data nowadays, 
better compression methods are constantly sought to preserve higher fidelity under the desired bitrate constraint. On the other hand, a better compression method should also be computationally efficient, which enables high-throughput processing.

With the rapid advancement of deep learning, end-to-end optimized lossy image compression has made remarkable progress over the last few years \cite{toderici2015variable, balle2018variational, chenglearned, 7999241,
minnen2018joint,he2021checkerboard,minnen2020channelwise,he2022elic,qian2022entroformer,koyuncu2022contextformer, liu2023learned, jiang2023mlic, jiang2023mlic++,
hu2021learning, qian2022learning, gao2021neural, wu2021learned, xieEnhancedInvertibleEncoding2021, theis2022lossy, guo2021causal,baoNonlinearTransformsLearned2023,wangEnsembleLearningBasedRateDistortion2021}. Currently, state-of-the-art (SOTA) learning-based methods~\cite{gao2021neural, wu2021learned,minnen2020channelwise, he2022elic, zou2022devil, koyuncu2022contextformer} have achieved superior performance to that of traditional compression methods, such as VVC~\cite{brossOverviewVersatileVideo2021}, which implies a promising future for learned image compression. Most learned image compression methods are based on an auto-encoder framework. The original input image is mapped to latent representations using nonlinear transforms and then quantized to discrete values. 
The discrete latent variables are encoded by arithmetic coding tools with the probability distribution estimated by an entropy model. As shown in~\cite{balle2018variational}, the final bitrate to losslessly compress the latent variables is the cross entropy between the prior and estimated distribution of the latent variables. 
It indicates that more accurate estimation leads to fewer bits required to compress an image, 
which motivates researchers to design more powerful entropy models.
\IEEEpubidadjcol

Recently, in light of the success of Transformer in natural language processing~\cite{vaswani2017attention}, many researchers~\cite{zou2022devil, bai2022endtoend,qian2022entroformer,koyuncu2022contextformer} have adopted transformer architectures in image compression field. Transformer-based entropy models, such as Entroformer~\cite{qian2022entroformer} and Contextformer~\cite{koyuncu2022contextformer}, have gained prominence due to their superior ability to capture long-range dependencies in probability distribution estimation compared to CNNs.
Initially, the latent variables are rearranged into a token sequence. Then, it is proposed to aggregate contexts with a stack of masked transformer decoder blocks, yielding large receptive fields, as well as content-adaptive transform. 
However, previous transformer-based entropy models still have limitations. 
As illustrated in Table \ref{tab:schemes-compare}, Entroformer~\cite{qian2022entroformer} only considers spatial correlation and suffers from a spatial serial decoding process. Its parallel variant Entroformer-P reduces decoding steps to 2, but the compression performance drops significantly.
Compared to Entroformer, Contextformer~\cite{koyuncu2022contextformer} further takes channel-wise correlation into account, but suffers from a even more complex spatial-channel serial coding process.
Moreover, both Entroformer and Contextformer overlook redundant context computation in the attention layer during inference.
All these factors make previous transformer-based entropy models time-consuming.

\begin{table}[t]
    \centering
    \caption{Comparing GroupedMixer with other entropy models: ``Conv.'' and ``Trans.'' refer to the use of convolutional modules and transformer modules in entropy model. ``Sp.'' and ``ch.'' denote spatial and channel contexts. ``G'' is the group count, and ``Dec.'' represents average encoding/decoding time for a $768\times 512$ image, ``Share'' denotes that building different autoregressive steps and mining different types of correlations with a set of shared parameters.
    }
    \small
    \setlength{\tabcolsep}{2.8pt} 
    \renewcommand{\arraystretch}{1.2}
    \begin{tabular}{lcccccc}
    \toprule
         Method & Type & Context & $G$ & Dec.(<1s) & Share \\
    \midrule
         Autoregressive~\cite{minnen2018joint} & Conv. & sp. & $HW$ & \XSolidBrush &\CheckmarkBold \\
         \midrule 
         \multicolumn{5}{l}{\textit{Group-based Entropy Model}}\\
         Checkerboard~\cite{he2021checkerboard} & Conv.& sp. & \textbf{2} & \CheckmarkBold &\CheckmarkBold \\
         ChARM~\cite{minnen2020channelwise} & Conv. & ch. & \textbf{10} & \CheckmarkBold &\XSolidBrush  \\ 
         ELIC~\cite{he2022elic} & Conv.& \textbf{sp.+ch.} & \textbf{10} & \CheckmarkBold &\XSolidBrush\\
         TCM~\cite{liu2023learned} &\textbf{Trans.}& \textbf{sp.+ch.} & \textbf{10} & \CheckmarkBold &\XSolidBrush \\
        MLIC~\cite{jiang2023mlic, jiang2023mlic++} &\textbf{Trans.}& \textbf{sp.+ch.} & \textbf{20} & \CheckmarkBold &\XSolidBrush \\
         \midrule 
         \multicolumn{5}{l}{\textit{Autoregressive Transformer}} \\
         Entroformer~\cite{qian2022entroformer}& \textbf{Trans.} & sp. & $HW$ & \XSolidBrush &\CheckmarkBold \\
         Entroformer-P~\cite{qian2022entroformer}& \textbf{Trans.}& sp. & \textbf{2} & \XSolidBrush &\CheckmarkBold \\
         Contextformer~\cite{koyuncu2022contextformer}& \textbf{Trans.}& \textbf{sp.+ch.} & $4HW$& \XSolidBrush &\CheckmarkBold \\
         \midrule 
          \multicolumn{5}{l}{\textit{Group-wise Autoregressive Transformer}} \\
         GroupedMixer& \textbf{Trans.}& \textbf{sp.+ch.} & \textbf{40} & \CheckmarkBold &\CheckmarkBold \\
         GroupedMixer-Fast& \textbf{Trans.}& \textbf{sp.+ch.} & \textbf{10} & \CheckmarkBold &\CheckmarkBold \\
    \bottomrule
    \end{tabular}
    \label{tab:schemes-compare}
\end{table}

In this paper, we introduce a novel transformer-based entropy model with group-wise token-mixers, dubbed GroupedMixer, to surmount the aforementioned challenges. 
In GroupedMixer, we first partition the latent variables into groups along spatial and channel dimensions. Then, we employ two group-wise token-mixers, \ie, inner-group and cross-group token-mixers to integrate contextual information within each group and across previously decoded groups respectively. Moreover, we carefully design position encodings for the proposed group-wise token-mixers to provide spatial-channel positional information.
Alternate arrangement of these two token-mixers enables global contextual reference. We compare our approach with previous entropy models~\cite{minnen2018joint,he2021checkerboard,minnen2020channelwise,he2022elic,qian2022entroformer,koyuncu2022contextformer, liu2023learned, jiang2023mlic, jiang2023mlic++} to distinguish it from various aspects in Table~\ref{tab:schemes-compare}. 
Firstly, our approach explores spatial-channel correlations existing in latent variables, as contrast to models focusing solely on spatial or channel dimensions~\cite{he2021checkerboard,minnen2020channelwise,qian2022entroformer}.
Secondly, our method employs a transformer architecture for global referencing instead of convolution~\cite{minnen2018joint,he2021checkerboard,minnen2020channelwise,he2022elic}, implemented with decomposed attention mechanism with lower computation rather than full attention~\cite{qian2022entroformer,koyuncu2022contextformer}.
Thirdly, relying on group-wise autoregression instead of pixel-wise autoregression~\cite{minnen2018joint, qian2022entroformer, koyuncu2022contextformer}, our approach only requires a constant number of inferences during coding. 
Fourthly, in contrast to previous group-based entropy models~\cite{minnen2020channelwise, he2022elic, liu2023learned, jiang2023mlic, jiang2023mlic++}, our method shares weights across different group predictions. This approach enables more effective modeling of complex correlations.
Additionally, we introduce context cache optimization to GroupedMixer to further accelerate the network inference, which caches attention activation values in cross-group token-mixer and avoids duplicated computation of contextual information. The context cache optimization not only enables acceleration of coding speed to under 1 second, but also effectively reduces multi-step fine-tuning time, which eliminates train-test gap and further boosts our performance.
Experimental results demonstrate that our proposed GroupedMixer achieves SOTA RD performance, while maintaining fast coding speed.

The contributions of our work are summarized as follows:
\begin{itemize}
    \item We present an entropy model named as GroupedMixer, a group-wise autoregressive transformer, which globally explores spatial-channel redundancies simultaneously. To lower the computation complexity, we propose to decompose the causal self-attention into two group-wise token-mixers, namely, inner-group and cross-group token-mixers.
    \item To further accelerate the network inference, we introduce context cache optimization to GroupedMixer, which enables faster coding speed and also can be applied to reduce multi-step fine-tuning time.
    \item With the help of the proposed entropy model and accelerating technique, we can achieve SOTA compression performance with fast compression speed. Specifically, we achieve 17.84\%, 19.77\% and 22.56\% BD-Rate savings on Kodak, CLIC'21 Test and Tecnick datasets when compared to VVC.
\end{itemize}

The rest of this paper is organized as follows. Section~\ref{sec:relatedwork} reviews related works. Section~\ref{sec:method} presents the proposed transformer-based entropy model. Experimental results are shown in Section~\ref{sec:experiments} and concluding remarks are given in Section~\ref{sec:conclusion}.

\section{Related Work\label{sec:relatedwork}}

\subsection{Learned Image Compression}

Learning-based image compression has witnessed remarkable advances in recent years, including lossless image compression \cite{bai2022deep,kingmaBitSwapRecursiveBitsBack2019, townsend2019practical, zhangIFlowNumericallyInvertible}, near-lossless image compression \cite{Bai_2021_CVPR} and lossy image compression \cite{toderici2015variable,balle2018variational,minnen2018joint}.
Learning-based lossy image compression originated from the work of Toderici \etal~\cite{toderici2015variable}.
Ball\'e \etal~presented an end-to-end optimized framework based on CNN in~\cite{balle2018variational}, which can be formulated as variational auto-encoder (VAE)~\cite{kingma2013autoencoding}. Jiang \etal~\cite{7999241} innovate with a convolutional neural network based end-to-end compression framework. Guo \etal~\cite{guo2021causal} present a novel causal context model based on CNNs to capture spatial-channel correlations in latent spaces. Wang \etal~\cite{wangEnsembleLearningBasedRateDistortion2021} explore rate-distortion optimization in image compression through ensemble learning. Wu \etal~\cite{wu2021learned} introduce a block-based hybrid image compression framework, enhancing efficiency and reducing memory issues through novel prediction and post-processing techniques. Finally, Bao \etal~\cite{baoNonlinearTransformsLearned2023} reveal a novel perspective on nonlinear transforms in learned image compression, proposing a nonlinear modulation-like transform inspired by communication techniques.  In contrast to the papers listed, this study introduces GroupedMixer, a novel transformer-based entropy model for learned image compression. This model leverages group-wise autoregression and decomposes full self-attention into more efficient group-wise token-mixers. Additionally, it incorporates context cache optimization for faster inference. 
Our compression model adapts VAE framework as in~\cite{balle2018variational}. In this framework, given an input natural image $\bm{x}$ in distribution $p_x$, an analysis transform $g_a$ is applied to decorrelate it into the compact latent representations $\bm{y}$. The latent variables $\bm{y}$ are then quantized to discrete values $\bm{\hat{y}}$ and further losslessly compressed using entropy coding techniques with an estimated distribution $p_{\hat{\bm{y}}} (\hat{\bm{y}})$. At the decoder side, the received quantized latent variables $\bm{\hat{y}}$ are recovered to a reconstruction $\bm{\hat{x}}$ using a synthesis transform $g_s$. We formulate the above process as:
\begin{equation}
\begin{aligned}
\bm{y} & = g_a(\bm{x};\bm{\phi}), \\
\bm{\hat{y}} &  = Q(\bm{y}), \\
\bm{\hat{x}} &  = g_s(\bm{\hat{y}}; \bm{\theta}),
\end{aligned}
\label{equ1}
\end{equation}
where $\phi$ and $\theta$ denote the parameters of analysis transform and synthesis transform respectively.
In learned image compression, the network can be trained with a loss which is a trade-off between rate and distortion as follows:
\begin{equation}
    \label{eq:rd_loss}
    \begin{aligned}
        R+\lambda \cdot D & =\mathbb{E}_{\boldsymbol{x} \sim p_{\boldsymbol{x}}}\left[-\log _{2} p_{\hat{\boldsymbol{y}} }(\hat{\boldsymbol{y}})\right] 
 +\lambda \cdot \mathbb{E}_{\boldsymbol{x} \sim p_{\boldsymbol{x}}}[d(\boldsymbol{x}, \hat{\boldsymbol{x}})], 
    \end{aligned}
\end{equation}
where the first term denotes the rate of the latent variables, the second term measures the reconstruction quality with a predefined distortion metric $d$ (\eg, PSNR and MS-SSIM), and $\lambda$ is a Lagrange multiplier which controls the RD trade-off during training.

\subsection{Entropy Model}
Entropy model plays a crucial role in estimating the prior distribution of quantized latent variables $p_{\hat{\bm{y}}}(\hat{\bm{y}})$. According to~\cite{balle2018variational}, more accurate estimation leads to lower bitrate.
To capture general structural information in latent variables, Ball\'e \etal~\cite{balle2018variational} introduced hyperprior $\bm{z}$, which is additionally encoded as side information to estimate distribution of $\hat{\bm{y}}$.
Minnen \etal~\cite{minnen2018joint} proposed a context model based on masked convolution layer to exploit spatial redundancies in latent variables. 
Thus, equipped with hyperprior and context model, rate term in \eqref{eq:rd_loss} can be reformulated as:
\begin{equation}
    \begin{aligned}
        R = \mathbb{E}_{\boldsymbol{x} \sim p_{\boldsymbol{x}}}\left[\sum_{i} -\log _{2} p_{\hat{y}_i \mid \hat{\bm{z}}, \hat{\bm{y}}_{<i} }(\hat{y}_i \mid \hat{\bm{z}}, \hat{\bm{y}}_{<i}; \bm{\psi})\right],
    \end{aligned}
\end{equation}
where $\psi$ is the parameter of entropy model.
Since then, many works have been proposed to improve the context model. We summarize prior efforts related to ours as following two main categories: 
\subsubsection{Global Prediction}
Several studies have expanded the context model's receptive field beyond local areas. 
Guo \etal~\cite{guo2021causal} split latent representations along the channel dimension, using global prediction for the second group based on similarity calculations with the first group's elements.
Entroformer~\cite{qian2022entroformer} developed a transformer-based context model leveraging global spatial correlations and accelerated it using a checkerboard-like parallelization. Contextformer~\cite{koyuncu2022contextformer} also considered channel-wise correlations. In these transformer-based methods, latent variables are initially rearranged into token sequences. The introduction of stacked masked transformer decoder blocks in place of masked convolution layers offers global receptive fields and context-adaptive transformation. Stacking transformer blocks enhances feature collection and improves prediction accuracy.

\begin{figure*}[h]
    \centering
    \includegraphics[scale=0.42]{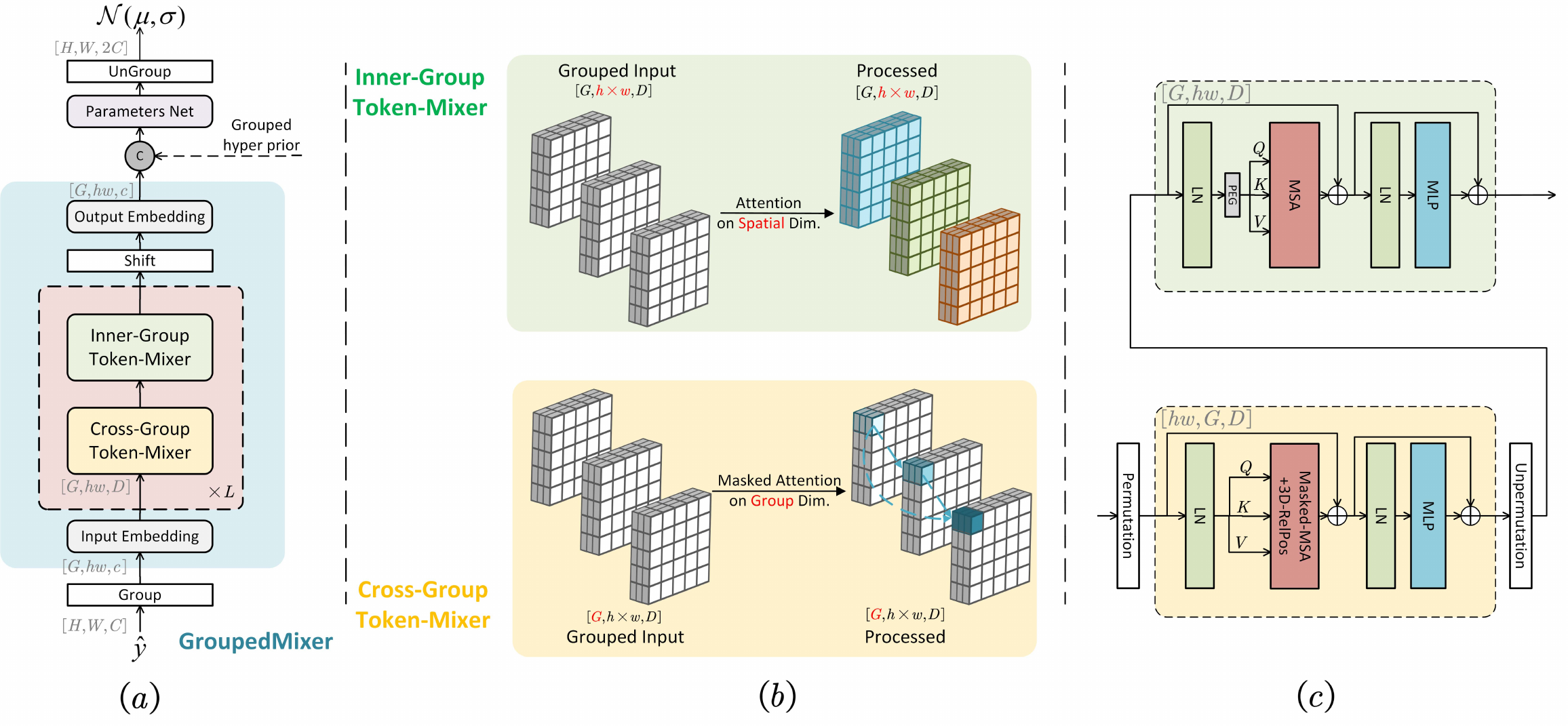}
    \caption{
    (a) \textbf{GroupedMixer overview\label{subfig:overview_a}}. 
    Preprocessed latent representations $\bm{\hat{y}}$ are then passed through GroupedMixer modules to aggregate group-wise context, and are finally projected as distribution parameters.
    (b) \textbf{Illustration of token-mixers\label{subfig:overview_b}}.
    Cross-group token-mixer mixes the information between previously decoded groups, while inner-group token-mixer mixes the information within groups. 
    (c) \textbf{Detailed network architectures\label{subfig:overview_c}} of two token-mixers, where MSA represents multi-head self-attention, and PEG denotes position embedding generator.
    } 
    \label{fig:overview}
\end{figure*}

\subsubsection{Group-based Autoregression}
Spatial autoregression, traditionally requiring serial coding, is accelerated through group-based methods. These methods divide latent variables into k groups across spatial-channel dimensions for parallelized coding within groups, leading to k-step inference. He \etal~\cite{he2021checkerboard} spatially split latent variables into two groups, employing checkerboard-like masked convolution for a two-step parallel model. ChARM~\cite{minnen2020channelwise} segmented latent variables into k channel slices, introducing a channel-wise context model. ELIC~\cite{he2022elic} developed an unevenly grouped channel-conditional model combined with a checkerboard spatial context, effectively creating spatial-channel group-based models. However, these methods, often reliant on convolutional layers, are limited to local receptive fields and struggle with long-range dependencies. Recent studies have concentrated on enhancing group-wise autoregression approaches with transformer modules. TCM~\cite{liu2023learned} improved the predictive power of the ChARM network by integrating Swin-Transformer blocks. MLIC~\cite{jiang2023mlic} was designed to capture channel-wise features using CNNs and to address both local and global spatial correlations through masked transformer modules. An advancement over MLIC~\cite{jiang2023mlic}, MLIC++~\cite{jiang2023mlic++} further improved upon this by altering the network architecture and incorporating linear attention computation.

It's important to note that previous group-based models, including TCM~\cite{liu2023learned} and MLIC~\cite{jiang2023mlic, jiang2023mlic++}, used various sets of CNNs and transformer modules to model each conditional distribution in the autoregressive model, differing fundamentally from our approach. Our entropy model, GroupedMixer, is a group-wise autoregressive transformer. This means that GroupedMixer shares weights across predictions of different groups by exploiting common predictive patterns. This shared-weight approach not only streamlines the model but also allows GroupedMixer to more effectively model complex correlations.

\begin{figure}[t]
 \centering
 \includegraphics[scale=0.53]{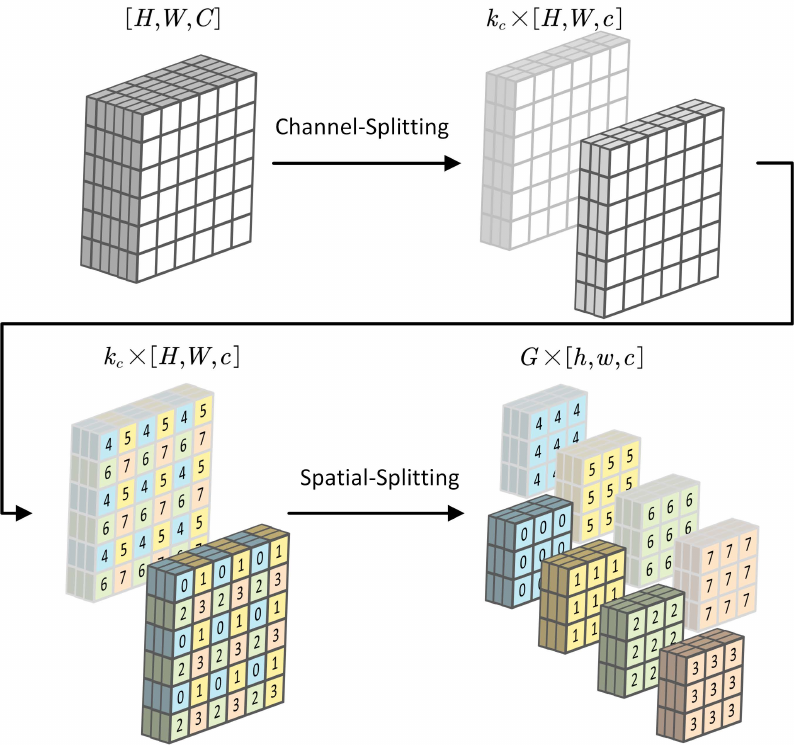}
 \caption{Grouping scheme for modeling spatial-channel context. The latent representations are separated along channel and spatial dimensions into $G=k_c\cdot k_h \cdot k_w$ groups sequentially, and number indicates order of autoregression. In this figure, we use $(k_c,k_h,k_w)=(2,2,2)$ as an example. }
\label{fig:split_schemes}
\end{figure}

\section{Proposed Method\label{sec:method}}  
Given an input image $\bm{x}$, we employ an analysis transform $g_a$ to decorrelate it into the latent variables $\bm{y}$. The latent variables $\bm{y}$ are then quantized to discrete values $\bm{\hat{y}}$ and losslessly compressed using entropy coding techniques, such as arithmetic coding, based on an estimated distribution. At the decoder side, the received quantized latent variables $\bm{\hat{y}}$ are reconstructed to $\bm{\hat{x}}$ using a synthesis transform $g_s$. The latent variables are assumed to follow a single Gaussian distribution $\mathcal{N}(\bm{\mu}, \bm{\sigma})$ and the parameters $\bm{\mu}, \bm{\sigma}$  are jointly estimated with a hyperprior model and our proposed entropy model, GroupedMixer.
In the following, we overview the network architecture of the proposed GroupedMixer and specify the details of the group-wise token-mixers.

\subsection{Overview of GroupedMixer}

We first introduce the network architecture of the proposed entropy model GroupedMixer, which is illustrated in Figure~\ref{fig:overview}a. Given a quantized latent representation $\hat{\boldsymbol{y}} \in \mathbb{R}^{H\times W\times C}$, we first separate it into a constant number of groups along both spatial and channel dimensions. 
We evenly split the latent variables into $k_c$ slices along the channel dimension, and then each channel slice is spatially segmented into $k_h \times k_w$ groups using a spatial splitting scheme, as illustrated in Figure\;\ref{fig:split_schemes}.
We define the group partition function $f$ and its inverse ungroup function $f^{-1}$ as follows:
\begin{eqnarray}
    \label{eq:group_partition}
    \begin{aligned}
    f&: \mathbb{R}^{H\times W\times C} \rightarrow \mathbb{R}^{G\times hw \times c},  \\
    f^{-1}&: \mathbb{R}^{G\times hw \times c} \rightarrow \mathbb{R}^{H\times W\times C},
    \end{aligned}
\end{eqnarray}
where the number of group is $G = k_c \cdot k_h \cdot k_w$ and the size of each group is $ hw\times c, h=\frac{H}{k_h}, w=\frac{W}{k_w}, c=\frac{C}{k_c}$.
We introduce two kinds of spatial splitting schemes, enabling 2 and 4 steps in the spatial autoregression respectively. For the 2-step splitting scheme, we adopt~\cite{he2021checkerboard} and partition the latent variables into a checkerboard pattern, which implies that $k_h=1, k_w=2$. For the 4-step splitting scheme, we select $k_h=2, k_w=2$ and divide latent variables as shown in Figure \ref{fig:split_schemes}.
With these spatial splitting schemes, our context model can leverage bidirectional spatial contextual information. 
In the following sections, the notation $k_c \times k_h k_w$ is consistently employed to represent the method of latent variables division. This denotes that the division process initially splits along the channel into $k_c$ slices, followed by a $k_h k_w$-step spatial division to further group each slice.
In our study, two configurations of latent variables division are employed: $10 \times 4$ for the base model and $5 \times 2$ for the fast model, respectively.

After the grouping operation, the latent variables $\hat{\boldsymbol{y}}$ are divided into groups $\hat{\boldsymbol{y}}_{\mathcal{G}_1}, \hat{\boldsymbol{y}}_{\mathcal{G}_2}, \cdots, \hat{\bm{y}}_{\mathcal{G}_G}$, where $\mathcal{G}$ denotes the set of all indices of symbols in the latent variables and it satisfies $\mathcal{G} = \mathcal{G}_1 \cup \mathcal{G}_2 \cdots \cup \mathcal{G}_G$. 
Each group of representation satisfies that $y_{\mathcal{G}_i} \in \mathbb{R}^{hw \times c}$. 
Our entropy model factorizes the joint probability of the latent variables as a group-wise autoregression as follows:
\begin{equation}
    p(\hat{\boldsymbol{y}} \mid \hat{\boldsymbol{z}}) = \prod_{i=1}^{G} p(\hat{\boldsymbol{y}}_{\mathcal{G}_i} \mid \hat{\boldsymbol{y}}_{\mathcal{G}_{<i}}, \hat{\boldsymbol{z}}).
\end{equation}
Compared with previous group-based methods \cite{minnen2020channelwise,he2022elic}, instead of modeling each conditional distribution with a separate network, we decouple the number of groups with network design and parameterize group-wise autoregressive model with a shared transformer architecture, GroupedMixer.

In GroupedMixer, instead of directly attending to all decoded values in previous groups, which is computationally intensive and usually impractical, we decompose the interactions with previous values into two token-mixers: cross-group and inner-group token-mixers,  to model global dependency across groups and within groups respectively. 
Specifically, we first map the $c$-channel representation into $D$ dimensions using an input embedding layer. Then the grouped representation is fed to our transformer model, which consists of multiple GroupedMixer modules, as detailed in Figure~\ref{fig:overview}a. Each module comprises cross-group and inner-group token-mixers. Given the grouped and mapped input feature $X_1 \in \mathbb{R} ^{G\times hw \times D}$, a GroupedMixer module can be expressed as:
\begin{equation}
    \begin{aligned}
        &\bm{Y}_l= \text{Cross-Group Token-Mixer}_l (\bm{X}_l), \\
        &\bm{Z}_l= \text{Inner-Group Token-Mixer}_l (\bm{Y}_l), 
    \end{aligned}
\end{equation} 
where $\bm{Y}_l$ and $\bm{Z}_l$ are intermediate feature and output feature in l-th layer module. The whole GroupedMixer architecture is constructed by repeatedly stacking the inner-group and cross-group token-mixers.

\begin{figure}[t]
 \centering
 \includegraphics[scale=0.6]{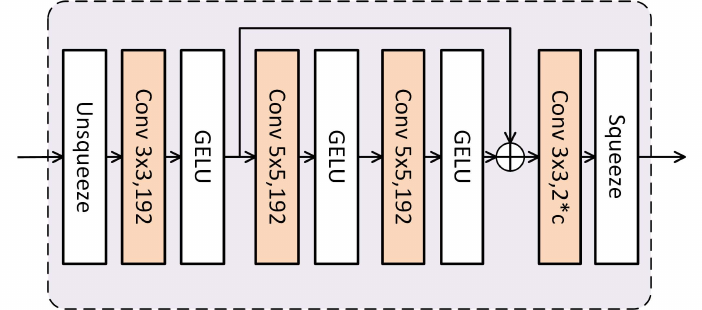}
 \caption{Detailed structure of Parameters Net. $c$ is the number of channels in each group.}
\label{fig:ep_net}
\end{figure}

The output feature $\bm{Z}_L \in \mathbb{R}^{G\times hw\times D}$ from the transformer blocks is shifted rightward along the second dimension, thus discarding the last group of features. A learnable parameter $\bm{\theta}_h \in \mathbb{R}^{hw \times 1 \times D}$ is then inserted at the first group's position, which is formulated as:
\begin{equation}
    \bm{Z}_L^s = [\bm{\theta}_h, \bm{Z}_L(:, :G-1, :)],
\end{equation}
where $[\cdot]$ denotes concatenation operation along the second dimension.
The motivation behind the shift operation is to ensure that the values of the current group is not accessible while predicting its parameters, aligning with the objectives of previous studies~\cite{qian2022entroformer,koyuncu2022contextformer}. 
Then the shifted feature undergoes an output embedding layer to map it back to the original dimension $c$ from $D$.
The representation is further concatenated with grouped hyper prior along the channel dimension and sent to parameter net. In parameter net, we first unsqueeze features to recover spatial dimension, from $\mathbb{R}^{G\times hw \times c}$ to $\mathbb{R}^{G\times c\times h\times w}$, and then aggregate hyperprior and context information within the same group using a stack of multiple convolution layers, and finally squeeze feature to original shape, as detailed in Figure~\ref{fig:ep_net}. 
At the end, the predicted parameters are rearranged to the same shape as the latent variables $\hat{\bm{y}}$ using ungroup operation $f^{-1}$ and split into $\bm{\mu},\bm{\sigma}$.

\subsection{Group-wise Token Mixers with Acceleration\label{sec:groupedmixer_module}}
We next specify the details of two token-mixers, and further propose a context cache optimization scheme to accelerate the network inference:

\subsubsection{Inner-Group Token-Mixer}
Inner-group token-mixer is designed to model the dependency within each group. 
As shown in the top subfigure of Figure~\ref{fig:overview}b, given the grouped representation $\boldsymbol{Y} \in \mathbb{R}^{G\times hw \times D}$, we apply an attention process along the spatial dimension and share the weights across the group dimension. Specifically, $\boldsymbol{Y}$ is sent to a multi-head self-attention module to mix information along the spatial dimension and produce the output feature $\boldsymbol{Y}^o \in \mathbb{R}^{G\times hw\times D}$. 
In the attention mechanism, we necessitate position embedding to provide the following attention process with the information of the current position. Considering discrepancy of image resolution in training and testing phases, instead of using diamond-shape position embedding in~\cite{qian2022entroformer}, we introduce a position embedding generator (PEG) \cite{chuConditionalPositionalEncodings2022} using depth-wise convolution, which introduces inductive bias of translation invariance in training phase. We define our position encoding scheme as follows:
\begin{equation}
    \bm{Y}^P = \operatorname{DWConv}(\bm{Y}) + \bm{Y},
\end{equation}
where the weights are initialized with zeros. We empirically find that the proposed position encoding scheme generalizes better on large resolution datasets in Section~\ref{subsubsec:position_embedding}.
We then linearly project position-aware input $\bm{Y}^P$ to queries, keys and values, and then split them into $m$ representations of each head, yielding $\boldsymbol{Q}_i^P, \boldsymbol{K}_i^P, \boldsymbol{V}_i^P\in \mathbb{R}^{G\times hw\times d_h},  \forall i=1, \cdots, m$, where $d_h$ is the number of head dimensions.
For spatial dimension,
Then we can formulate multi-head attention process as follows:
\begin{equation}
    \label{eq:attention}
    \begin{aligned}
        \boldsymbol{A}_{i} & = \sigma\left(\frac{\boldsymbol{Q}_i^P \boldsymbol{K}_i^{P \top}}{\sqrt{d_h} }\right) \boldsymbol{V}_{i}^P, \\
        \boldsymbol{Y}^{o} & = \boldsymbol{W}\left(\boldsymbol{A}_{1} \oplus \cdots \oplus \boldsymbol{A}_{m}\right) + \boldsymbol{Y},
    \end{aligned}
\end{equation}
where $\sigma$ denotes the softmax function.
Then $\boldsymbol {Y}^{o}$ is sent to a feed-forward network to mix information along the channel dimension using two linear layers with expansion ratio $4$ and produce the output feature $\boldsymbol {Z}\in\mathbb {R}^{G\times hw\times D}$. In this way, different tokens in each group can be adequately mixed along spatial and channel dimensions for generating output features.

\begin{figure}[t]
 \centering
 \includegraphics[scale=0.48]{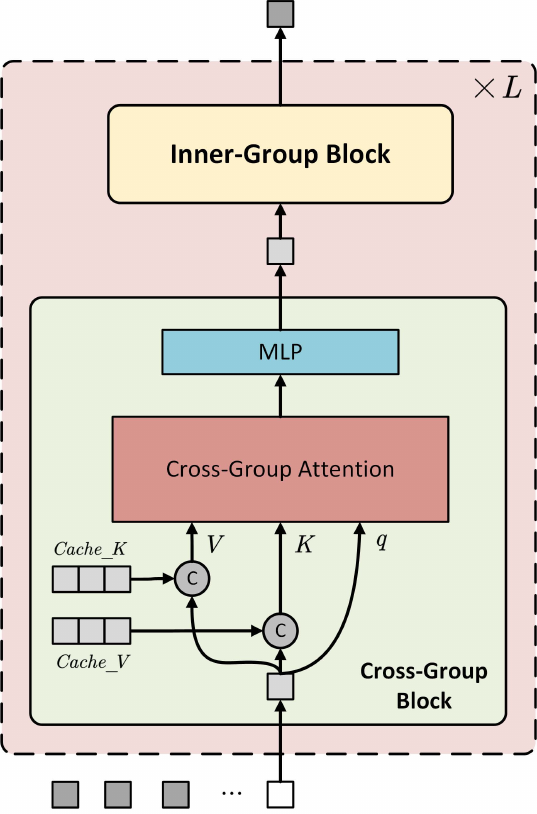}
 \caption{Context cache optimization at inference time. Each block denotes a group of attention activation values. 
 }
\label{fig:attn_cache}
\end{figure}

\subsubsection{Cross-Group Token-Mixer}
Cross-group token-mixer exchanges information across different groups, and thus makes the model to aggregate global spatial-channel information from previous decoded values. As shown in the bottom subfigure of Figure \ref{fig:overview}b,
the cross-group token-mixer is implemented by applying \textit{masked} attention mechanism on the group dimension. For implementation, given the input feature $\boldsymbol{X} \in \mathbb{R}^{G\times hw \times D}$, we first permute the group and spatial dimensions, and get the rearranged intermediate feature $\boldsymbol{X}^{r} \in \mathbb{R}^{hw\times G\times D}$.
The input feature $\boldsymbol{X}^{r}$ undergoes a masked attention process to integrate information across the group dimension while maintaining the same weights on the spatial dimension.
We reformulate \eqref{eq:attention} as:
\begin{equation}
        \boldsymbol{A}_{i} = \sigma\left(\frac{\boldsymbol{Q}_i \boldsymbol{K}_i^{\top} + \boldsymbol{Q}_i \boldsymbol{P}^{C}}{\sqrt{d_h} } + \boldsymbol{M} \right) \boldsymbol{V}_{i}
\end{equation}
where $\bm{P}^C \in \mathbb{R}^{G\times G \times d}$ denotes relative position embedding, and $\boldsymbol{M} \in \mathbb{R}^{G\times G}$ signifies the mask matrix, in which the lower triangle is zero and the rest is $-\infty$.
To encode position information for distinguishing each group, consider the group dimension as a 3D space with size $(k_c, k_h, k_w)$. Each group index $i$ is thus assigned a 3D coordinate $(x_i, y_i, z_i)$. For indexes $u, v$, the relative position vector $\mathbf{r}_{u,v}$ is computed as $\mathbf{r}_{u,v} = (x_u-x_v, y_u-y_v, z_u-z_v)$, where each component of the vector $(x_u - x_v, y_u - y_v, z_u - z_v)$ falls within the range $[-k_h+1, k_h-1], [-k_w+1, k_w-1], [-k_c+1, k_c-1]$ respectively, and is then mapped to a positive scalar $s$:
\begin{equation}
    \begin{aligned}
        s = (z_u-z_v + k_c-1)&\cdot(2 k_h+1)\cdot(2k_w+1) \\ 
        + (x_u-x_v + &k_h-1 ) \cdot (2k_w +1) \\ 
        + y_u-y_v &+ k_w-1.  \\
    \end{aligned}
\end{equation}
The relative position embedding $\bm{P}^C_{u,v}\in\mathbb{R}^{d}$ between group $u$ and $v$ can be indexed as:
\begin{equation}
    \bm{P}^C_{u, v} = \bm{W}_{s,:},
\end{equation}
where $\bm{W} \in \mathbb{R}^{n_p \times d}$ is a 2D learnable parameter and initialized with truncated normal distribution and $n_p = (2k_c-1)\cdot(2k_h-1)\cdot(2k_w-1)$.
Similar to inner-group token-mixer, the output feature is sent to a MLP module to further mix information along the channel dimension and produce the output feature $\boldsymbol {Y}\in\mathbb {R}^{G\times hw\times D}$.
Note that this cross-group process implies that information is aggregated across accessible groups but restricted to the same spatial location, as illustrated in the bottom subfigure of Figure~\ref{fig:overview}b. By alternately stacking inner-group and cross-group token-mixers, the model can adequately mix information along the spatial, group and channel dimensions for capturing global information.

\subsubsection{Context Cache Optimization for Acceleration\label{sec:accelerating}}

At inference time, we estimate the distribution of the latent variables using group-wise autoregression, which requires $G$ times inference of the context model. However, the inference speed remains a challenge due to the large model size and intensive computation involved in the autoregressive process.
One critical factor overlooked by previous works \cite{qian2022entroformer,koyuncu2022contextformer} is the repeated context computation during inference, which implies that the network inference inputs a gradually enlarging context from $0$ to $G-1$ groups of tokens at each of the G steps.
Inspired by the attention cache optimization to expedite sequence generation~\cite{yanFastSeqMakeSequence2021} in NLP, we propose context cache optimization for faster group-wise autoregressive distribution prediction, as illustrated in Figure~\ref{fig:attn_cache}. 
Considering that only cross-group token-mixer performs interactions with previously decoded groups, the corresponding activations need to be cached and are utilized for being attended in later inference stages.
Therefore, we cache activations including each group of keys and values in every cross-group token-mixers at each inference time. Before performing attention process, key and value produced by the current group are concatenated with cached activation values. Note that mask is canceled in attention because all concatenated decoded keys and values can be accessed at current step.
The self-attention with context cache optimization in the cross-group token-mixer at step $t$ is implemented in \eqref{eq:cache_attn}:
\begin{equation}
    \label{eq:cache_attn}
    \begin{aligned}
        \boldsymbol{Q}^t_i &= \boldsymbol{X}^t_i \boldsymbol{W}^Q , \\
        \boldsymbol{K}^t_i &= [\text{Cache\_K}_{t-1}, \boldsymbol{X}^t_i \boldsymbol{W}^K], \\
        \boldsymbol{V}^t_i &= [\text{Cache\_V}_{t-1}, \boldsymbol{X}^t_i \boldsymbol{W}^V] , \\
        \boldsymbol{A}^t_i &= \sigma(\frac{\boldsymbol{Q}^t_i \boldsymbol{K}^{t\top}_i + \boldsymbol{Q}^t_i\boldsymbol{P}_c^{\top}}{\sqrt{D}}) \boldsymbol{V}^{t}_i,
    \end{aligned}
\end{equation}
where concatenated keys and values satisfy $\bm {Q}_i^{t}\in\mathbb {R}^{hw\times 1\times D},\bm {K}_i^{t}\in\mathbb {R}^{hw\times t\times D},\bm {V}_i^{t}\in\mathbb {R}^{hw\times t\times D}$. 
Our context cache optimization diverges from~\cite{yanFastSeqMakeSequence2021} in the following aspects:
(1) \textit{ Unlike the approach in \cite{yanFastSeqMakeSequence2021} where attention results from all blocks are stored, we cache attention activation values specific to group-wise autoregression in cross-group token-mixers. } 
(2) \textit{Our strategy operates at the group level, caching and concatenating intermediate activation values \(\mathbb{R}^{hw \times 1 \times D}\) for a group of tokens rather than individual tokens.} 
These modifications not only tailor the attention activation cache concept to our GroupedMixer's architecture but also represent a novel application of this strategy.
Equipped with context cache optimization, we only need to feed a single group of latent variables $\hat{\boldsymbol {y}}_{\mathcal{G}_t}$ to GroupedMixer each  time instead of all decoded groups $\hat{\boldsymbol {y}}_{\mathcal{G}_{<t}}$. Thus, the inference process of GroupedMixer is significantly accelerated, leading to faster image coding.

\subsection{Analysis\label{sec:analysis}}
\subsubsection{Computation Complexity\label{sec:accelerating}}
In the GroupedMixer module, instead of directly applying full self-attention process on all tokens which entails computational complexity:
\begin{equation}
\mathcal{O}(Attn) = 4(Ghw)d_h^2 + 2(G h w)^2 d_h.
\end{equation}
We decompose the global attention into two parts: inner-group and cross-group token-mixers. The computational complexity of the inner-group token-mixer is $\mathcal{O}(4(hw)d_h^2 + 2(hw)^2d_h )$, and the computational complexity of the cross-group token-mixer is $\mathcal{O}(4Gd_h^2+2G^2d_h)$. The overall complexity of this decomposition is reduced to follows:
\begin{equation}
    \mathcal{O}(GM) = 4(hw+G)d_h^2 + 2((hw)^2+G^2)d_h,
\end{equation}
where the major computation and memory cost comes from the self-attention process in inner-group token-mixer. This decomposition also enables us to scale to larger $G$, which is configured to $10, 40$ in our implementation.
To better understand the computational complexity, consider an image size of $768\times 512$, downscaled to $32\times 48$ latent codes and then grouped into 40 groups ($10\times 4$ configuration), retrieving $h=16, w=24$. The overall computational complexity of our GroupedMixer module is:
\begin{equation}
    \begin{aligned}
        \mathcal{O}(GM) &=4(16\times 24+40)d_h^2 + 2((16\times 24)^2 + 40^2)d_h \\
        &= 1,696 d_h^2 + 298,112 d_h 
    \end{aligned}
\end{equation}
This demonstrates the efficiency of our method compared to directly applying full attention process, whose complexity is 
\begin{equation}
    \begin{aligned}
        \mathcal{O}(Attn) &= 4(40\times16\times24)d_h^2 + 2(40\times 16\times 24)^2 d_h \\
        &=61,440 d_h^2 + 471,859,200  d_h   
    \end{aligned}
\end{equation}

\subsubsection{Insights for Enhanced Performance\label{sec:insights}}
In this section, we delve deeper into the underlying reasons for the enhanced compression performance, increased coding speed, and reduced parameter count achieved by GroupedMixer.
Contrasting with existing group-based entropy models, such as ChARM~\cite{minnen2020channelwise}, ELIC~\cite{he2022elic}, which employ convolutional layers to model dependencies, our approach leverages the transformer architecture to capture long-range dependencies, outperforming conventional methods in terms of performance. Moreover, our model significantly reduces the number of parameters by utilizing a shared transformer architecture, as opposed to modeling each group with distinct network parameters as in ChARM~\cite{minnen2020channelwise}, ELIC~\cite{he2022elic}, TCM~\cite{liu2023learned}, MLIC~\cite{jiang2023mlic, jiang2023mlic++}. This design choice enables more effctive modeling complex correlations with the same amount of aprameters and enhances its scalability to larger group sizes. 
Different from previous transformer-based entropy models Entroformer~\cite{qian2022entroformer}, Contextformer~\cite{koyuncu2022contextformer} that utilize pixel-wise autoregression and full self-attention across all tokens, our model employs group-wise autoregression and innovatively splits the attention mechanism into inner-group and cross-group token-mixers.
These designs decrease the number of autoregression steps and further lower computational complexity of each inference. Alongside with context cache optimization, they speed up inference, making a notable advancement over Entroformer~\cite{qian2022entroformer}, Contextformer~\cite{koyuncu2022contextformer}.
These also facilitate efficient multi-step fine-tuning, addressing train-test mismatch for improved performance, as shown in Section~\ref{subsubsec:msft}. 
Our model addresses redundancies across the spatial and channel dimensions, rather than solely the spatial dimension as in Entroformer~\cite{qian2022entroformer}.
Due to reducing computation complexity, our approach can segment latent variables into more slices across the channel dimension. This, coupled with carefully designed position embeddings, results in enhanced performance compared to Contextformer~\cite{koyuncu2022contextformer}. Consequently, our model attains superior performance to earlier transformer-based entropy models with even less complex analysis and synthesis transform network, coupled with the benefits of faster inference speeds. This represents a significant advancement in the field of efficient learned image compression.

\begin{figure*}[h!]
    \captionsetup[subfigure]{labelformat=empty}
      \centering
     \begin{subfigure}{.46\linewidth}
        \centering\includegraphics[width=1.\linewidth]{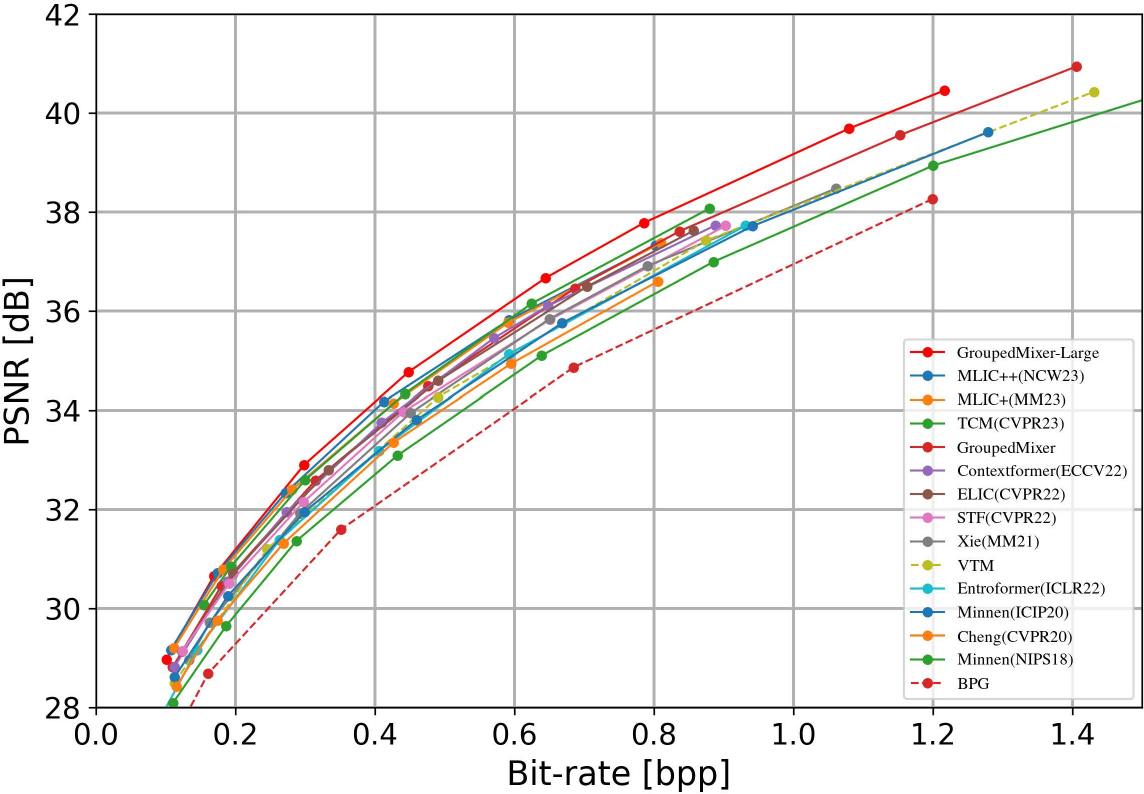}
      \end{subfigure}
      \begin{subfigure}{.46\linewidth}
        \centering\includegraphics[width=1.\linewidth]{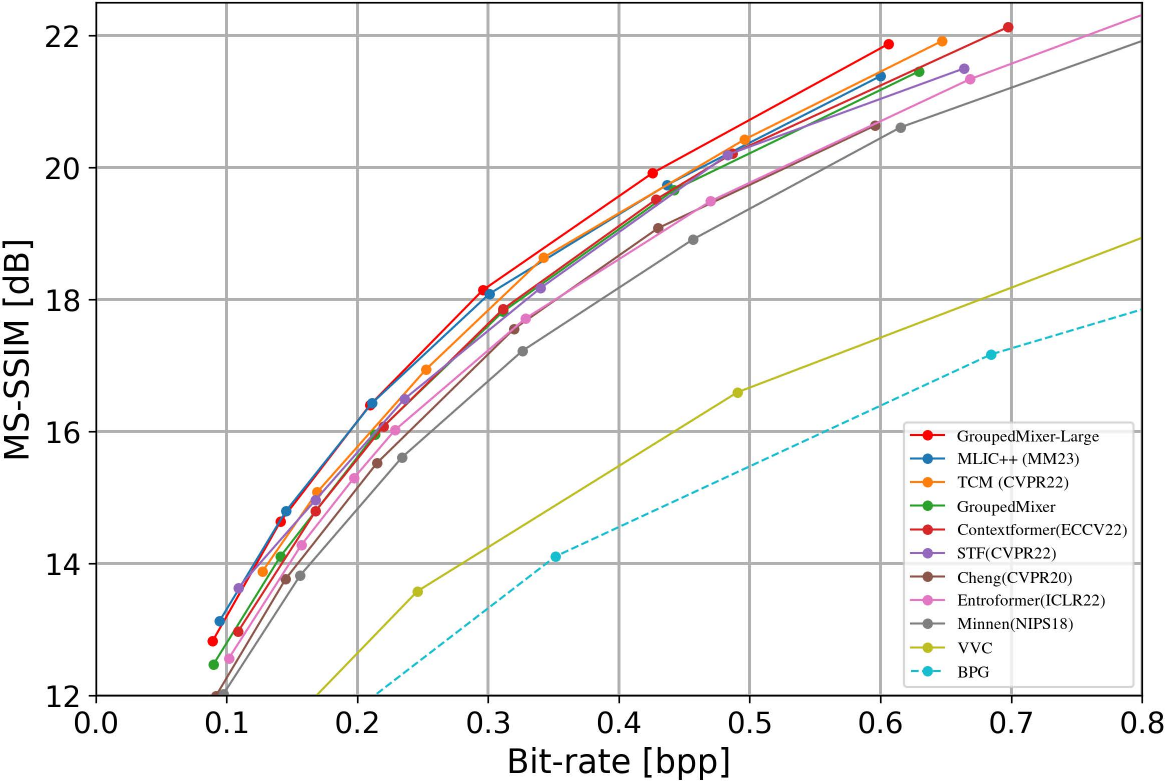}
      \end{subfigure}
      \caption{
        Performance evaluation on the Kodak dataset. }
        \label{fig:kodak}
    \end{figure*}
    
\begin{figure*}[htbp]
    \vspace{-0.4cm}
    \centering
    \begin{minipage}[t]{0.48\textwidth}
    \centering
    \includegraphics[width=8.2cm]{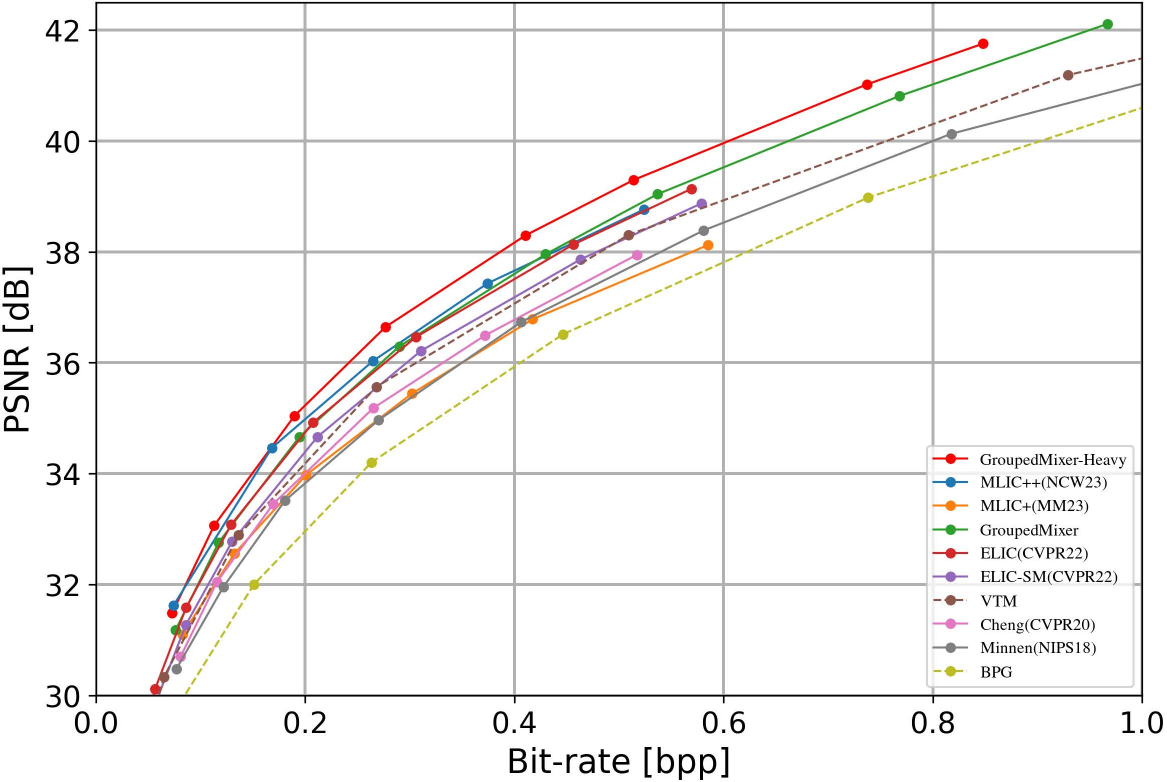}
    \caption{
        Performance evaluation on the CLIC'21 Test dataset.}
    \label{fig:clic}
    \end{minipage}
    \begin{minipage}[t]{0.48\textwidth}
    \centering
    \includegraphics[width=8.2cm]{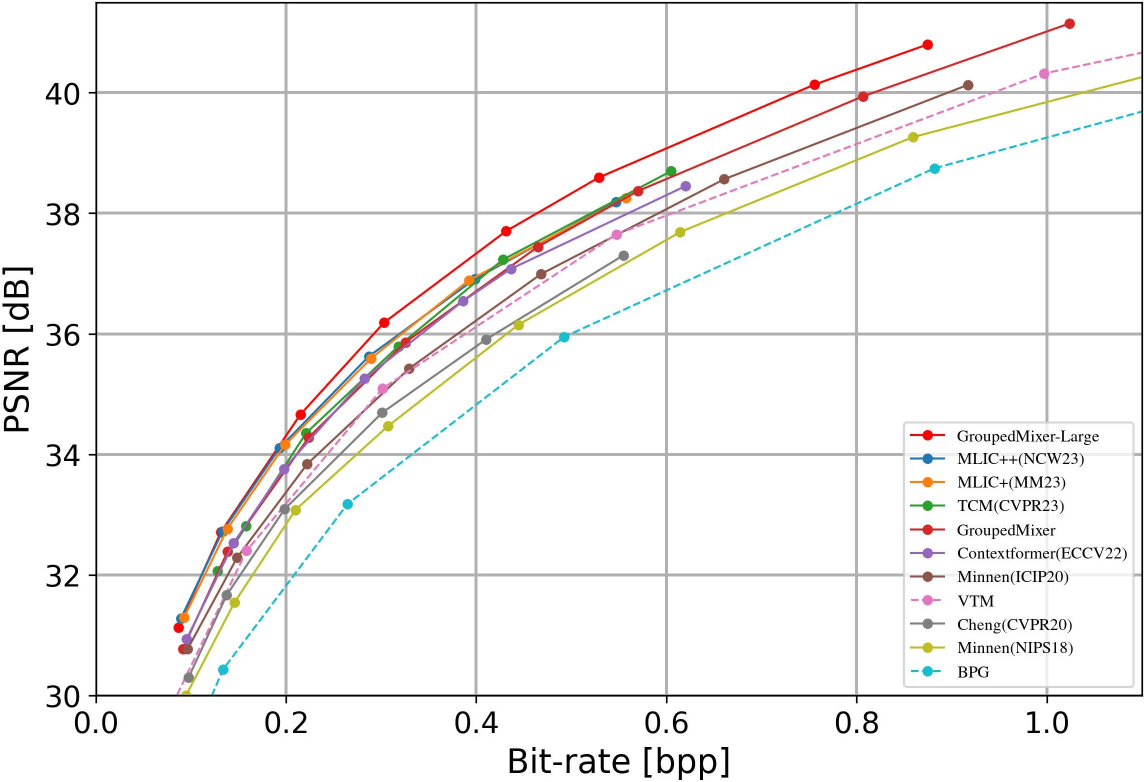}
    \caption{
        Performance evaluation on the Tecnick dataset.}
    \label{fig:tecnick}
    \end{minipage}
\end{figure*}
    
\begin{table*}[!t]\setlength{\tabcolsep}{10pt}\renewcommand{\arraystretch}{1.2}
    \centering
    \caption{Evaluation of different methods on the Kodak dataset. \textbf{``BD-Rate''} is computed with VTM as anchor method. \textbf{``Tot. Enc.''} and \textbf{``Tot. Dec.''} indicate total encoding and decoding times in seconds (\textbf{s}), \textbf{``T. Dec.''} covers network transformation time during decoding, and \textbf{``E. Dec.''} includes entropy coding time (hyper prior, context model, arithmetic coding). Total parameter (\textbf{``Tot. Param.''}) are in millions (\textbf{M}), with \textbf{``T. Param.''} and \textbf{``E. Param.''} for transformation and entropy model parameters, respectively. $^*$ marks results from original papers, $^\dag$ denotes our reimplemented model tests. ``CCO'' refers to context cache optimization.}

    \begin{tabular}{lcccccccccc}
    \toprule
    \multirow{2}{*}{\textbf{Model}}  & \multirow{2}{*}{\textbf{BD-Rate(\%)}} &  \multicolumn{4}{c}{\textbf{Latency(s)}} &\multicolumn{3}{c}{\textbf{Params(M)}} \\
         && \textbf{Tot. Enc.}& \textbf{Tot. Dec.} & \textbf{T. Dec.} &\textbf{E. Dec.}  &\textbf{Tot. Param.} &\textbf{T. Param.} &\textbf{E. Param.}\\
         \midrule
         VVC~\cite{brossOverviewVersatileVideo2021} &  0.00 & 129.21 & 0.14 & / &/ &/ &/ &/\\ 
         \midrule
         Minnen~\cite{minnen2018joint} & 10.90 & 1.98& 4.60 &6e-3 &4.59 & 25.5 &7.0 &18.5\\
         Cheng~\cite{chenglearned} & 5.44 & 1.98& 4.69 &0.03 &4.66 & 29.6 &19.0 &10.6\\
         Xie~\cite{xieEnhancedInvertibleEncoding2021}& -0.78 & 2.93 &  6.00 &1.92 &4.08  & 50.0 &39.3 &10.7\\
         \midrule 
         \multicolumn{5}{l}{\textit{Group-based Entropy Model}} \\
         ChARM~\cite{minnen2020channelwise} & 1.38 & 0.08$^\dag$ & 0.09$^\dag$ & 6e-3$^\dag$ &0.08$^\dag$ &126.7$^\dag$& 7.0$^\dag$ & 119.7$^\dag$ \\
         STF~\cite{zou2022devil}  & -4.31 & 0.14 & 0.13 &0.03 &0.10 & 99.9 &13.8 &86.1\\
         ELIC-SM~\cite{he2022elic} & -1.07 & 0.06$^\dag$ & 0.08$^\dag$ &6e-3 $^\dag$&0.07$^\dag$& 29.8$^\dag$ &7.6$^\dag$&22.2$^\dag$\\
         ELIC~\cite{he2022elic} & -7.24 & 0.07$^\dag$ & 0.09$^\dag$  &0.02$^\dag$&0.07$^\dag$& 36.9$^\dag$ &14.6$^\dag$&22.2$^\dag$\\
         TCM~\cite{liu2023learned}& -11.74 & 0.16 & 0.15 &0.07 &0.08 & 76.6 &28.9 & 47.7\\
         MLIC+~\cite{jiang2023mlic}& -13.11 & / &  /& /& / & / & / & / \\
         MLIC++~\cite{jiang2023mlic++}& -15.07 & 0.16 & 0.18 &0.02 &0.16 & 116.7 & 16.4 &100.3  \\
         \midrule 
         \multicolumn{5}{l}{\textit{Autoregressive Transformer}} \\
         Entroformer~\cite{qian2022entroformer}& 2.73 & 3.86 & 121.19   &0.03 &121.16& 45.0 &7.6&37.4 \\
         Entroformer-P~\cite{qian2022entroformer}& 5.50 & 4.28 & 7.95   &0.03 &7.93& 45.0 &7.6&37.4\\
         Contextformer~\cite{koyuncu2022contextformer} & -7.44 & 40$^*$ & 44$^*$ &/ &/ & 37.4$^*$ &17.5$^*$ &19.9$^*$\\
          \midrule 
          \multicolumn{5}{l}{\textit{Group-wise Autoregressive Transformer}} \\
         GroupedMixer& -8.31 & 0.27 & 0.26 &6e-3 &0.25 & 36.4 &7.6 &28.8\\
         \textit{- w/o CCO} & - & \textit{1.06} & \textit{1.01} &\textit{6e-3} &\textit{1.01} &- &-& - \\
         GroupedMixer-Fast& -6.03 & 0.07 & 0.06  &6e-3 &0.05 & 36.8 &7.6 &29.2 \\
         GroupedMixer-Large& -17.84 & 0.17 & 0.17 &0.07 &0.10 & 72.8 &28.9 &43.9\\
         \bottomrule
    \end{tabular}
    \label{tab:rd_speed_param}
\end{table*}

\begin{table}[!t]
    \centering
    \caption{Speed performance analysis of two models at varying resolutions. This comparison focuses on the latency metrics in milliseconds (\textbf{ms}) for both encoding and decoding processes across different image resolutions. }

    \small 
    \setlength{\tabcolsep}{3.0pt} 
    \renewcommand{\arraystretch}{1.2}
    \begin{tabular}{lccccc}
    \toprule
    \multirow{2}{*}{\textbf{Model}} & \multirow{2}{*}{\textbf{Size}} &  \multicolumn{4}{c}{\textbf{Latency(ms)}} \\
         & &  \textbf{Tot. Enc.}& \textbf{Tot. Dec.}& \textbf{T. Dec.} & \textbf{E. Dec.} \\
         \midrule
         \multirow{4}{*}{GM} 
         & $256\times 256$ & 242.40&  241.15 & 1.46 & 239.69 \\ 
         & $512\times 512$ &256.52 &251.34 &4.27 & 247.07\\
         & $1024\times 1024$ &363.20 &356.63 &15.50 &341.12\\
         & $2048\times 2048$ &1814.28 &1755.38 &59.31 &1696.07\\
         \midrule
         \multirow{4}{*}{GM-Fast} 
         & $256\times 256$ & 60.90  & 59.39 & 1.71 &57.68 \\ 
         & $512\times 512$ & 68.67 &63.26 &4.43 &58.82\\
         & $1024\times 1024$ &172.44 &158.91 &15.64 &143.28\\
         & $2048\times 2048$ &1218.46 & 1147.26 &59.23&1088.03\\
         \bottomrule
    \end{tabular}
    \label{tab:speed_analysis}
\end{table}

\section{Experiments\label{sec:experiments}}

\subsection{Settings \label{subsec:training_details}}

\subsubsection{Implementation Details}
As our main contribution is to design a novel entropy model, we directly adopt the transform network design in~\cite{he2022elic, liu2023learned}.
We share the same analysis transform $g_a$, synthesis transform $g_s$, hyper analysis transform $h_a$ and hyper synthesis transform $h_s$ with ELIC-SM~\cite{he2022elic} and TCM~\cite{liu2023learned}.
We set the number of channels $N=192$ and $M=320$ for all models. For architecture of transformer, we employ depth $L=6$, the embedding channels $d_e=384$ and the number of heads $m=12$ as our default configuration. 
We initialize the weights of GroupedMixer with a truncated normal distribution as used in ViT~\cite{dosovitskiy2022image}. 
In our study, for rapid comparison, we employ a transform network identical to that in ELIC-SM and opt for the $10\times 4$ partitioning method as the foundational variant (GroupedMixer). We also use the $5\times 2$ method for a faster variant, termed GroupedMixer-Fast. To present the best performance on compression and speed, we substitute the transform network in GroupedMixer-Fast with a mixed Transformer-CNN architecture, as proposed in TCM~\cite{liu2023learned}. This modified version is designated as GroupedMixer-Large.
For training, we use the Adam~\cite{kingma2014adam} optimizer with $\beta_1 = 0.9, \beta_2=0.999$.
All models are trained for over 2M iterations. 
Gradient clipping technique is also employed for stable training and the clip value is set to 1.0.
We optimize our models with rate-distortion loss as Equation~\eqref{eq:rd_loss}. During training, a Lagrange multiplier $\lambda$ balances the rate-distortion trade-off. When optimizing for MSE, we train different models with values of $\lambda \in \{ 4, 8, 16, 32, 75, 150, 300, 450, 900, 1500 \} \times 10^{-4}$, where $D=\operatorname{MSE}(\hat{x}, x)$. When optimizing for MS-SSIM, we utilize $\lambda \in \{2, 4, 8, 16, 32, 64 \}$, where $D=1-\operatorname{MS-SSIM}(\hat{x}, x)$.

Due to the repeated invocation of the entropy coder during decoding in the group-based entropy model, we have achieved speed acceleration by optimizing the calling interface, based on CompressAI~\cite{begaint2020compressai}. Specifically, we modified the data transfer method used when calling the entropy coder, changing it from list transfer to tensor transfer, which speeds up entropy coding process.
In our method, tensor reorganization is a crucial step to align data for subsequent processing. This is achieved through a rearrangement operation, which we implement using the \textit{einops.rearrange} function~\cite{rogozhnikov2022einops}. This function allows us to efficiently transform the multi-dimensional structure of our data tensors into the desired format. 
Our model is implemented based on Pytorch 2.0~\cite{paszke2019pytorch}, utilizing \textit{torch.nn.functional.scaled\_dot\_product\_attention} to accelerate the computation process of attention.

\subsubsection{Datasets}
Our compression model is trained on randomly selected $5\times10^{4}$ images from DIV2K~\cite{Agustsson_2017_CVPR_Workshops}, FLICKR 2K~\cite{hodosh2013framing}, ImageNet~\cite{ILSVRC15} with resolution larger than $480\times 480$ as our training set. 
We evaluate the rate-distortion performance on three publicly available datasets: Kodak dataset~\cite{kodak} with 24 images with a size of $768\times512$, Tecnick dataset \cite{asuni2014testimages} with 100 images with a size of $1200\times1200$, and CLIC'21 Test dataset \cite{clic} with 60 images with a size of 2k resolution. We measure the distortion of reconstruction using PSNR and MS-SSIM, and convert MS-SSIM to $-\log_{10}(1-\text{MS-SSIM})$ for clearer comparison. 
For CLIC'21 Test \cite{clic} and Tecnick \cite{asuni2014testimages}, we adhere to the testing protocol employed in CompressAI \cite{begaint2020compressai}, where the images are padded to multiples of $64$ beforehand, while the rate is derived from padded images and distortion is evaluated on images without padding.

\subsubsection{Three-Stage Training Scheme}
Following previous works~\cite{he2021checkerboard, minnen2018joint, he2022elic} and community discussion
\footnote{\url{https://groups.google.com/g/tensorflow-compression/c/LQtTAo6l26U/m/mxP-VWPdAgAJ}}
, we encode $\lceil \boldsymbol{y} - \boldsymbol{\mu} \rfloor$ into bitstream which enhances RD performance, and thus the decoder side receives the reconstructed latent variables $\hat{\bm{y}} = \lceil \boldsymbol{y} - \boldsymbol{\mu} \rfloor + \boldsymbol{\mu}$.
To train a group-based transformer, we adapt a three-stage training strategy. In the first stage, we employ one-pass training scheme for faster forward process, and feed differentiable estimation $\bm{y}+\mathcal{U}(-\frac{1}{2}, \frac{1}{2})$ to the context model, $\text{STE}(\bm{y})$ to the decoder, where $\mathcal{U}(-\frac{1}{2}, \frac{1}{2})$ is the uniform distribution and $\text{STE}(\cdot)$ is the straight-through estimator~\cite{theis2022lossy}. 
The original images are cropped to $256 \times 256$ patches before being fed into the network. Minibatches of 16 of these patches are used to update the network parameters and the learning rate is set to $1\times 10^{-4}$.
The first training stage lasts for 1.2M-step iterations.
From the second stage on, we enlarge the size of patches from 256 to 384 to train our transformer for better global context modeling ability and batch size is set to 8. The learning rate drops to $5\times 10^{-5}$ at the beginning of this stage, and drops to $1\times10^{-5}$ at 1.5M steps. The second stage lasts for 0.6M iterations.
In the third stage, we apply multi-step fine-tuning to close the train-test gap, which improves performance for group-based entropy models. Specifically, we conduct $G$ times inference of context model to retrieve real reconstruction  $\hat{\bm{y}} = \text{STE} (\boldsymbol{y} - \boldsymbol{\mu} ) + \boldsymbol{\mu}$, and feed it to both the context model and the decoder.
The learning rate starts at $4\times 10^{-5}$ and the learning rate is then divided by 2 when the evaluation loss reaches a plateau (we use a patience of $1\times 10^4$ iterations).
This stage finishes after at most 1M steps or when the learning rate is decayed below a certain threshold value, which is set to $1\times 10^{-6}$.
Thanks to the context cache optimization introduced in section~\ref{sec:accelerating}, our model can be trained with less computation and memory cost in this stage.

\subsection{Rate-Distortion Performance}

We compare our method with other learned image compression methods, including MLIC~\cite{jiang2023mlic, jiang2023mlic++}, TCM~\cite{liu2023learned}, Contextformer \cite{koyuncu2022contextformer}, Entroformer \cite{qian2022entroformer}, ELIC, ELIC-SM \cite{he2022elic}, STF \cite{zou2022devil}, ChARM \cite{minnen2020channelwise}, Xie \etal \cite{xieEnhancedInvertibleEncoding2021}, Cheng \etal \cite{chenglearned}, Minnen \etal \cite{minnen2018joint}, as well as some traditional codecs, like VVC intra (VTM 17.0) \cite{brossOverviewVersatileVideo2021} and BPG \cite{bellard2015bpg}. For ChARM~\cite{minnen2020channelwise}, we reproduce the method based on CompressAI~\cite{begaint2020compressai} for fair comparison. For ELIC, ELIC-SM~\cite{he2022elic}, we use publicly available unofficial implementation provided by~\cite{jiang2022unofficialelic}, which is based on CompressAI~\cite{begaint2020compressai}. For \cite{chenglearned} and \cite{minnen2018joint}, we evaluate the codes provided by CompressAI~\cite{begaint2020compressai}. The other results are obtained from their official repositories or emails with authors.

Figure \ref{fig:kodak} shows the rate-distortion performance on the Kodak dataset with the model optimized for MSE and MS-SSIM respectively. 
We also report BD-rates reduction of our method and state-of-the-art learned methods in Table \ref{tab:rd_speed_param}. The BD-rate is computed from the Rate-PSNR curve with VTM as the anchor method. 
This demonstrates that our methods outperform all listed methods in terms of PSNR and MS-SSIM.
Our GroupedMixer, GroupedMixer-Fast, GroupedMixer-Large reduce BD-rates by 8.31\%, 6.03\% and 17.81\% on Kodak dataset over VVC when measured in PSNR. 
From Figure~\ref{fig:kodak} we can see, the performance gain is more significant for higher bitrates. 
Compared to ELIC-SM~\cite{he2022elic}, our GroupedMixer, which employs the same transform network as ELIC-SM, achieves a 7.24\% BD-Rate saving. Additionally, it is capable of increasing the PSNR by up to approximately 0.5dB at the same bitrate. 
Compared with TCM~\cite{liu2023learned}, our GroupedMixer-Large, utilizing the same transform network as TCM, realizes a 6.07\% BD-Rate saving. This demonstrates that our proposed entropy model surpasses the convolution-based entropy model presented in ELIC~\cite{he2022elic} and the Swin-Transformer Attention based entropy model used in TCM~\cite{liu2023learned}. Compared with previous SOTA MLIC++~\cite{jiang2023mlic++}, our GroupedMixer-Large also achieves 2.74\% BD-Rate saving.
Compared with Entroformer, our GroupedMixer achieves 11.04\% BD-rate reduction. Even equipped with a lighter transform network than Contextformer~\cite{koyuncu2022contextformer} and ELIC~\cite{he2022elic}, our GroupedMixer still outperforms them by 1.07\% and 0.87\% BD-rate reduction respectively, especially at higher bitrate regions. 

The performance evaluation results on high-resolution image datasets CLIC'21 Test and Tecnick are respectively presented in Figure \ref{fig:clic} and Figure \ref{fig:tecnick}. As shown, our approach surpasses all other codecs.
The GroupedMixer-Large model demonstrates superior performance with BD-rate savings of 19.77\% and 22.56\% on the CLIC'21 Test and Tecnick datasets, respectively. This marks a significant improvement over the previous SOTA method, MLIC++~\cite{jiang2023mlic++}, which achieved BD-rate savings of 15.60\% and 18.36\% on these datasets.

\begin{figure*}[h]
    \centering
    \includegraphics[scale=1.1]{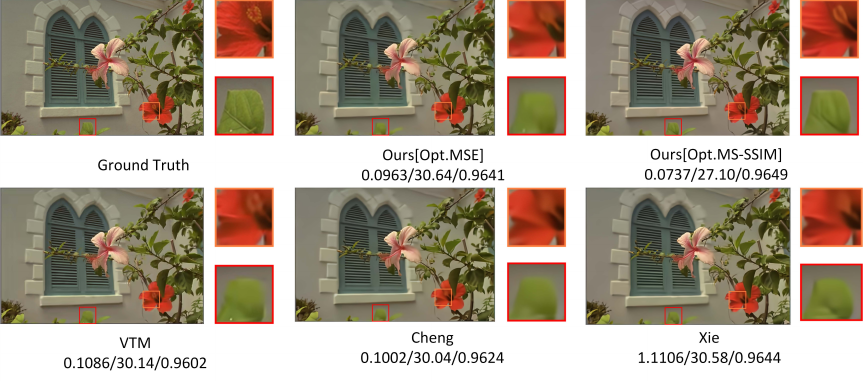
    }
    \caption{Visual comparisons of decompressed image of ``kodim07.png'' from Kodak dataset based on different methods. The subfugure is titled as ``bitrate/PSNR/MS-SSIM''.}
    \label{fig:vis_compare}
\end{figure*}

\subsection{Model Complexity} 
We report encoding/decoding latency and number of parameters in Table \ref{tab:rd_speed_param}.
For evaluation of model speed, we limit the accessible computation resources to one 3090 GPU card and AMD Ryzen 9 5900X 12-Core CPU. We use same testing protocol as ELIC~\cite{he2022elic} to measure the latency. Note that entropy coding details is not mentioned in ELIC~\cite{he2022elic}, so we directly use the publicly accessible reimplementation~\cite{jiang2022unofficialelic} to test coding latency. 
The table reports total and individual latencies for transform and entropy coding time respectively, averaged on all images in Kodak dataset. For computation of entropy coding time, we count inference time of hyperprior, context model and entropy coder.

As shown in Table~\ref{tab:rd_speed_param}, our method not only achieves SOTA RD performance but also excels in efficiency. When compared with other transformer-based entropy models, our approach demonstrates significantly faster speeds. For instance, our base model records average encoding and decoding times of 0.27s and 0.26s, respectively. This is 466 times faster than Entroformer~\cite{qian2022entroformer} and 169 times faster than Contextformer~\cite{koyuncu2022contextformer} in decoding. Our GroupedMixer-Fast variant further improves these speeds to 0.07s for encoding and 0.06s for decoding. It even surpasses ELIC-SM~\cite{he2022elic} in decoding speed while maintaining superior RD performance. These enhancements are primarily due to the context cache optimization and our refined entropy coding process, leading to a fivefold and twofold increase in decoding speed respectively.

Additionally, our experiments on varying image sizes reveal a direct correlation between resolution and coding time in GroupedMixer. For example, decoding time for a $256 \times 256$ image is 241.15 ms, increasing to 1755.38 ms for a $2048 \times 2048$ image. In contrast, the GroupedMixer-Fast variant shows a rise from 59.39 ms to 1147.26 ms for these resolutions. Notably, the latency growth rate accelerates with higher resolutions. Since the time complexity is predominantly influenced by the inner-group block complexity, which is proportional to the square of the number of latent variables.

In Table~\ref{tab:rd_speed_param}, we demonstrate that weight sharing across groups maintains a nearly constant parameter count, even as the number of groups increases. 
In fact, due to reductions in the input/output embedding layer size, the parameter count may decrease.
This represents an improvement over models like ChARM~\cite{minnen2020channelwise}, STF~\cite{zou2022devil}, TCM~\cite{liu2023learned} and MLIC~\cite{jiang2023mlic, jiang2023mlic++}, as our entropy model conserves parameters effectively, even with more groups.
Compared to previous models like ELIC~\cite{he2022elic}, Entroformer~\cite{qian2022entroformer}, and Contextformer~\cite{koyuncu2022contextformer}, GroupedMixer stands out for its superior RD performance, achieved with a comparable or lower total parameter count and a lighter transform network.
Detailed analysis reveals that our GroupedMixer has slightly more parameters than Contextformer~\cite{koyuncu2022contextformer} (18.3M vs. 15.9M), and the hyperprior network does carry more parameters (10.5M vs. 4.0M).
Although our context model and hyperprior network are slightly more parameter-intensive compared to Contextformer~\cite{koyuncu2022contextformer}, our strategic choices enable us to maintain leading encoding speeds, a more crucial aspect in practical applications.
We can observe that enhancing the transform network's capacity increases inference latency marginally, but significantly boosts performance. Our GroupedMixer-Large surpasses both TCM~\cite{liu2023learned} and MLIC~\cite{jiang2023mlic, jiang2023mlic++} in performance, even with fewer parameters. Specifically, our model saves 43.9M parameters compared to MLIC++~\cite{jiang2023mlic++} and achieves better compression performance, demonstrating the efficient modeling capabilities of our approach.

\newcommand{\supprogressivelw}{0.19\linewidth}
\begin{figure*}[htb]
    \centering
    \subfloat[$k=0$ \\ (0.007 / 14.72)]{
    \includegraphics[width=\supprogressivelw]{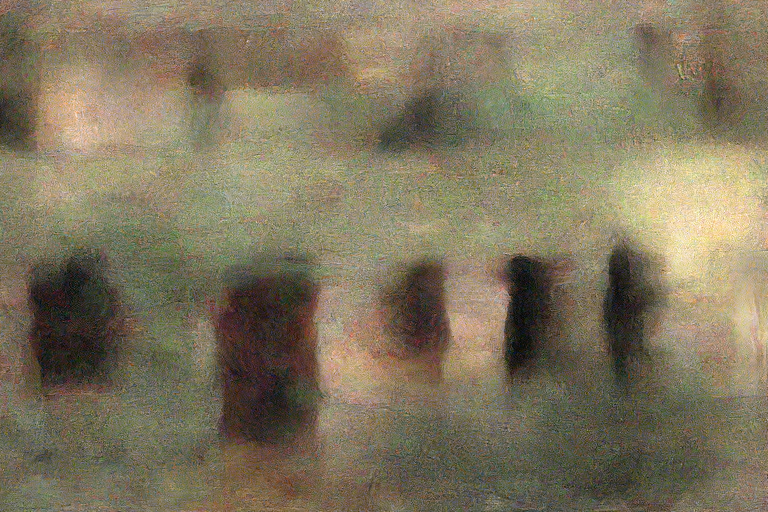}
    }
    \subfloat[$k=1$ \\ (0.069 / 17.78)]{
    \includegraphics[width=\supprogressivelw]{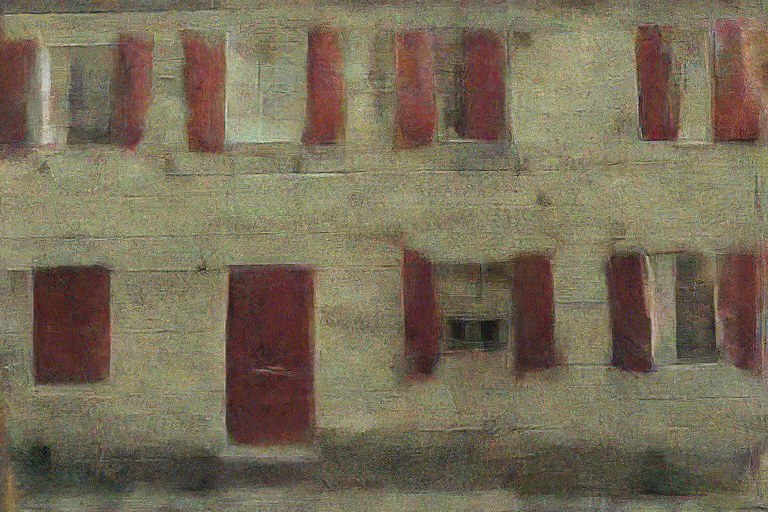}
    }
    \subfloat[$k=2$ \\ (0.125 / 18.64)]{
    \includegraphics[width=\supprogressivelw]{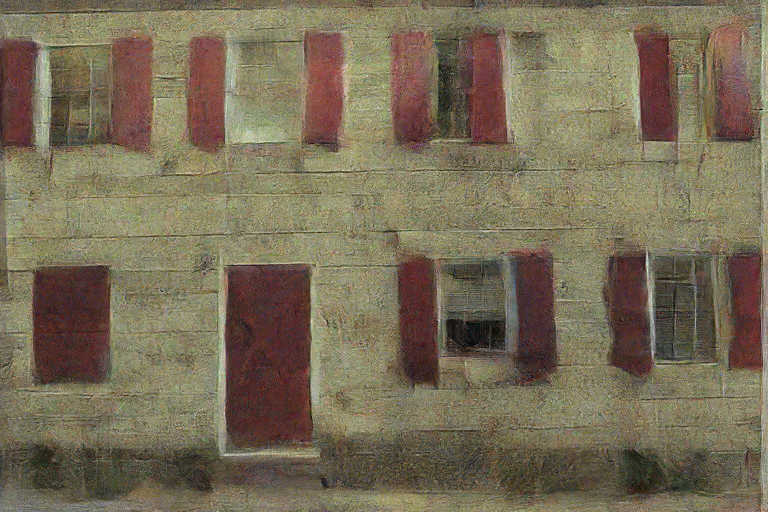}
    }
    \subfloat[$k=3$ \\ (0.235 / 19.97)]{
    \includegraphics[width=\supprogressivelw]{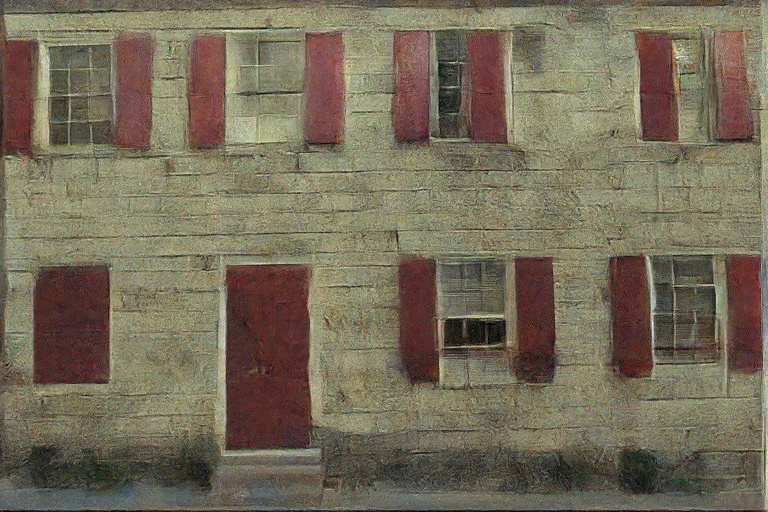}
    }
    \subfloat[$k=4$ \\ (0.429 / 21.19)]{
    \includegraphics[width=\supprogressivelw]{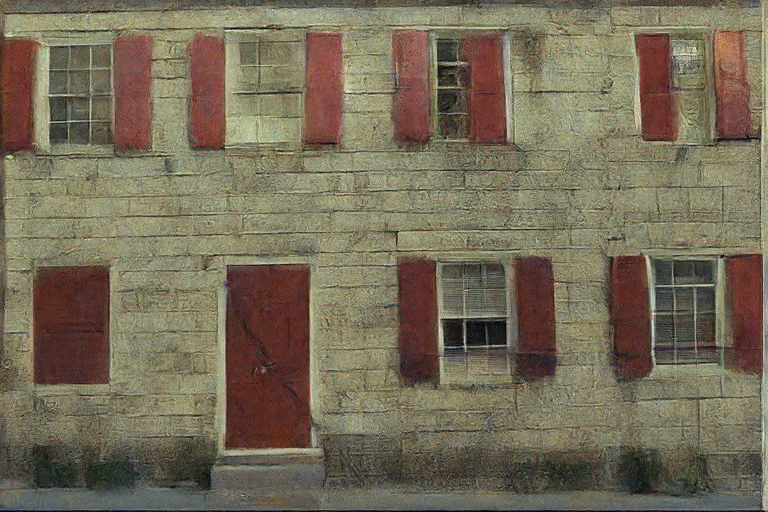}
    }
    \\
    \subfloat[$k=8$ \\ (0.601 / 22.21)]{
    \includegraphics[width=\supprogressivelw]{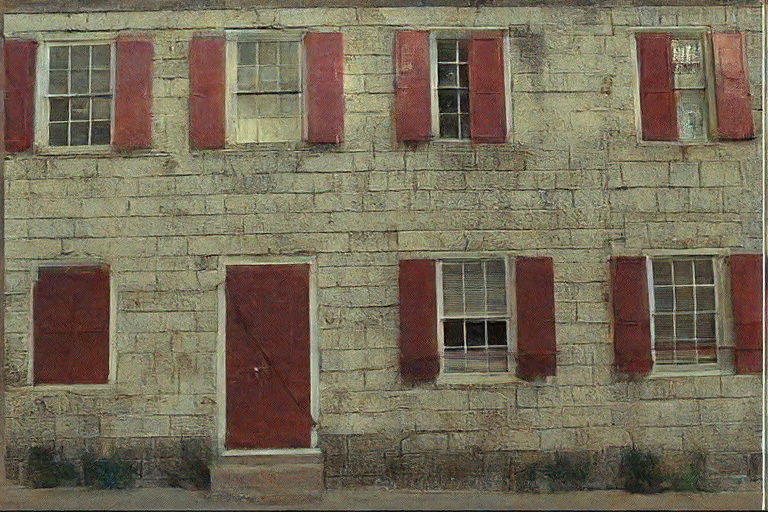}
    }
    \subfloat[$k=16$ \\ (0.756 /23.00)]{
    \includegraphics[width=\supprogressivelw]{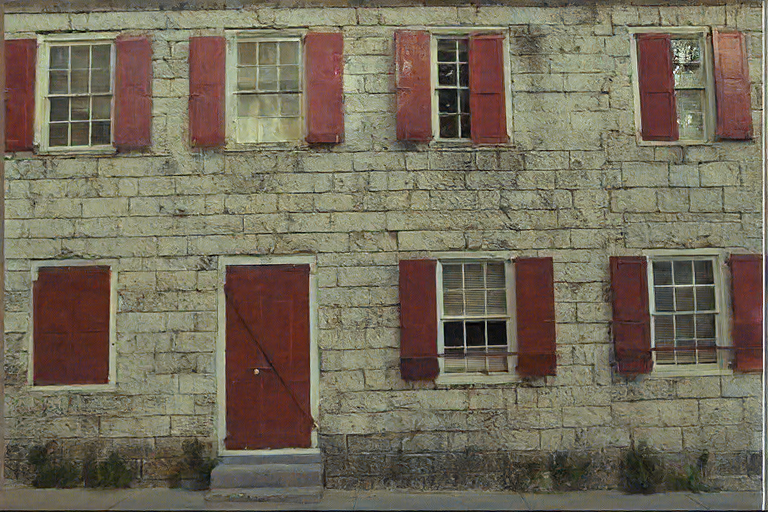}
    }
    \subfloat[$k=24$ \\ (1.009 / 26.16)]{
    \includegraphics[width=\supprogressivelw]{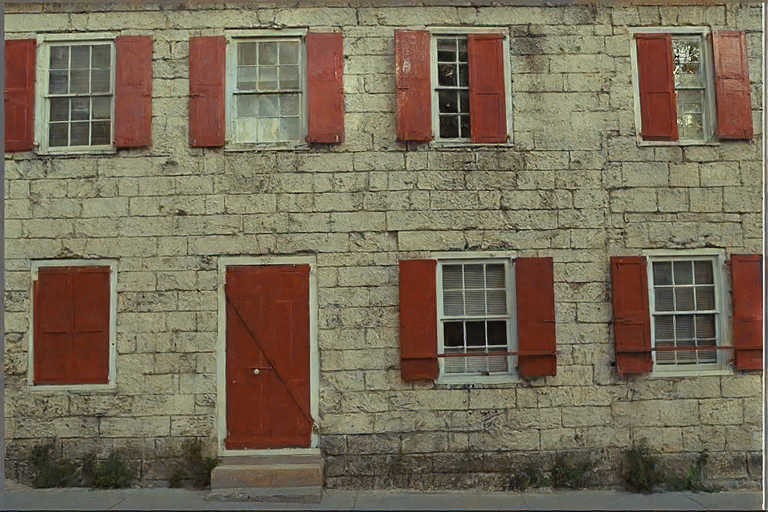}
    }
    \subfloat[$k=32$ \\ (1.207 / 29.04)]{
    \label{subfig:progressive-kodim01-32}
    \includegraphics[width=\supprogressivelw]{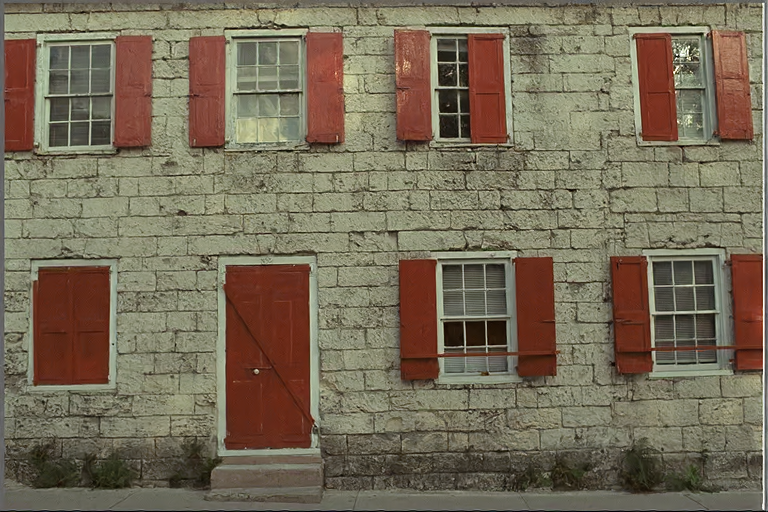}
    }
    \subfloat[$k=40$ \\ (1.412 / 36.21)]{
    \label{subfig:progressive-kodim01-40}
    \includegraphics[width=\supprogressivelw]{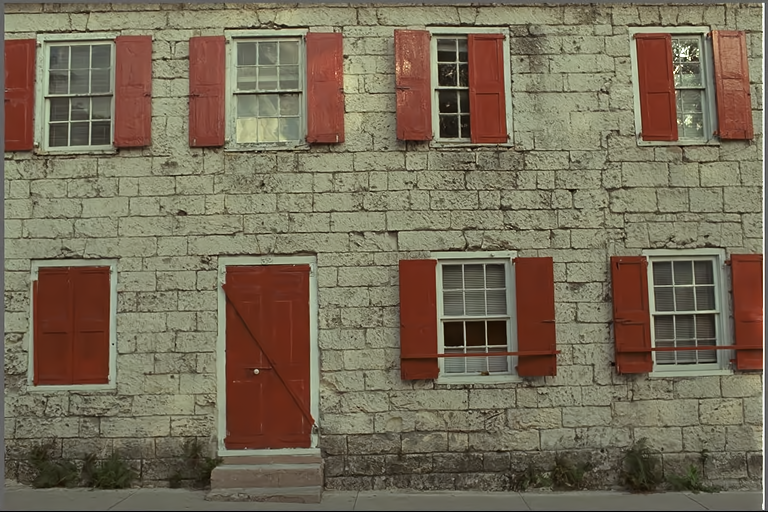}
    }
    \caption{Progressive decoding results of ``kodim01.png''. The subfugure is titled as number of received latent groups $k$ on decoder side and ``bitrate/PSNR''. It is recommended to zoom in and view these figures in full screen.}
    \label{fig:progressive-kodim01}
\end{figure*}

\subsection{Qualitative Results}
To illustrate that our method can generate visually appealing results, we visualize some reconstructed images for qualitative performance comparison.
We show some qualitative comparison of reconstructed images on the Kodak dataset in Figure \ref{fig:vis_compare}. This sample image ``kodim07.png'' is evaluated with an approximate bpp of 0.1. We can deduce from the figure that our method can preserve more details even at the lower bitrate, such as sharper texture of flower and leaves in ``kodim07.png''. It confirms that our MSE optimized method achieves better fidelity with less bitrate compared with the traditional codecs, \eg,  VTM~\cite{brossOverviewVersatileVideo2021} and BPG \cite{bellard2015bpg}, and surpasses the performance of other learned image compression method Cheng \etal \cite{qian2022entroformer} and Xie \etal \cite{xie2021enhanced}. Moreover, we can observe that our MS-SSIM optimized model achieve better visual quality than other codecs. 
We conclude that better RD performance result benefits from our proposed entropy model, which realizes more precise prediction and thus more bitrate savings.

\subsection{Progressive Decoding \label{subsec:progressive_Decoding}}
In literature, learned image compression with group-based entropy model can also be applied to implement progressive coding~\cite{minnen2020channelwise, he2022elic}. In this section, we also evaluate the ability of our method on progressive decoding. We directly adapt our model ($10\times 4$ variant) trained with MSE distortion and $\lambda=0.045$ for progressive decoding without any additional post-training. Assuming that the decoder side already receives bitstreams of $k$ latent groups, where $k=0$ denotes the case that no bitstream of $\hat{\bm{y}}$ but only $\hat{\bm{z}}$ is received. We first reconstruct $k$ groups of latent variables from bitstreams using entropy coding based on the predicted distribution. For the remaining $G-k$ groups, we view the context model as a generative model and sample latent variables from the predicted Gaussian distribution 
$p(\hat{\bm{y}}_{\mathcal{G}_i} \mid \hat{\bm{y}}_{\mathcal{G}_{<i}}, \hat{\bm{z}}; \psi)$ 
by ancestral sampling, 
where we first sample $\hat{\bm{y}}_{\mathcal{G}_{k+1}}$ from $p(\hat{\bm{y}}_{\mathcal{G}_{k+1}} \mid \hat{\bm{y}}_{\mathcal{G}_{<k+1}}, \hat{\bm{z}}; \psi)$, then sample $\hat{\bm{y}}_{\mathcal{G}_{k+2}}$ from $p(\hat{\bm{y}}_{\mathcal{G}_{k+2}} \mid \hat{\bm{y}}_{\mathcal{G}_{<k+2}},\hat{\bm{z}}; \psi)$, and so on. 
The progressive decoding results are shown in Figure~\ref{fig:progressive-kodim01}. Compared with other group-based methods \cite{he2021checkerboard,minnen2020channelwise, he2022elic}, our method scales to a larger number of groups and thus achieves finer progressive coding. From the figures, sampling only from the hyperprior ($k=0$) renders the image content as color blocks, indicating that the hyperprior encodes the blurry structure of the image. Note that when only one group of latent variables is decoded ($k=1$), the shape in sampled image for ``kodim01.png'' is still correlated and recognizable, which is partly due to global spatial-channel-wise context reference during sampling. As the received bitstream of latent variables $\hat{\bm{y}}$ increases, the sampled image gradually contains more precise structural information, such as the detail outline of windows and bricks in ``kodim01.png''. Then the image tone progressively changes, the details gradually enrich, both approaching the real image. From the above phenomenon, we can conclude that the initial latent groups tend to encode the structural information of the image, while the later latent groups contain the color and detail information. 

\subsection{Model Analysis \label{subsec:model_analysis}}
In this section, we conduct more experiments to further analyze GroupedMixer. In the following experiments, for rapid validation, we adopt simpler network architecture and training scheme. Specifically, we select the base configuration of GroupedMixer as $\{L=3, d_e = 384, m=12\}$ and use the variant $5\times2$ as default unless otherwise specified. 
For the training strategy, we first train our model with $
\lambda=0.045$ for 1M iterations with base learning rate $1\times 10^{-4}$. Then, we fine-tune the model with $\lambda=0.03, 0.015$ respectively, covering $\lambda = 0.015, 0.03, 0.045$ regions.

\begin{figure}[t]
    \begin{subfigure}{0.48\columnwidth}
	\centering
    \subfloat[Order Analysis]{
     \includegraphics[scale=0.36, clip]{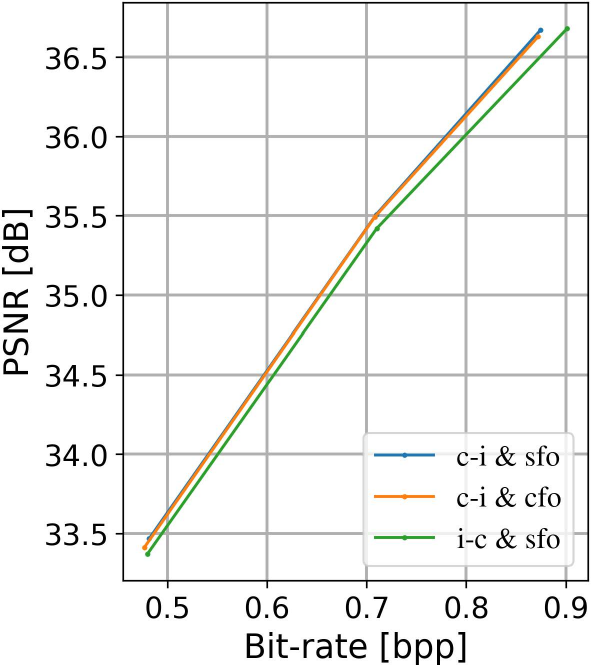}
    \label{subfig:ab_order}
    }
    \end{subfigure}
    \begin{subfigure}{0.48\columnwidth}
	\centering
    \subfloat[Position Embedding]{
     \includegraphics[scale=0.36, clip]{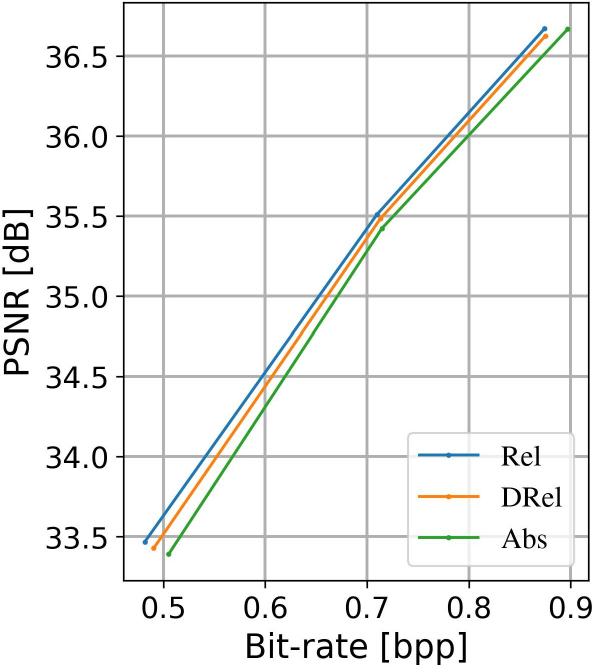}
    \label{subfig:ab_pos}
    }
    \end{subfigure}
    \begin{subfigure}{0.48\columnwidth}
	\centering
    \subfloat[PEG]{
     \includegraphics[scale=0.36, clip]{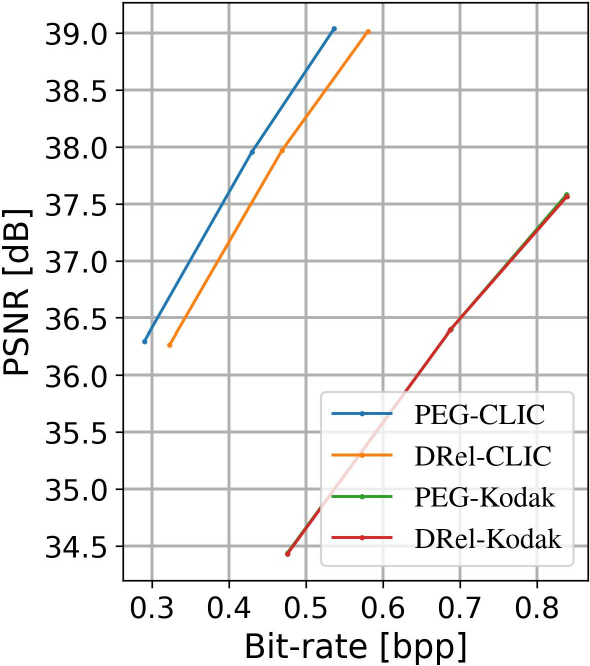}
    \label{subfig:ab_pos_peg}
    }
    \end{subfigure}
    \begin{subfigure}{0.48\columnwidth}
	\centering
    \subfloat[Multi-step Fine-tuning]{
     \includegraphics[scale=0.36, clip]{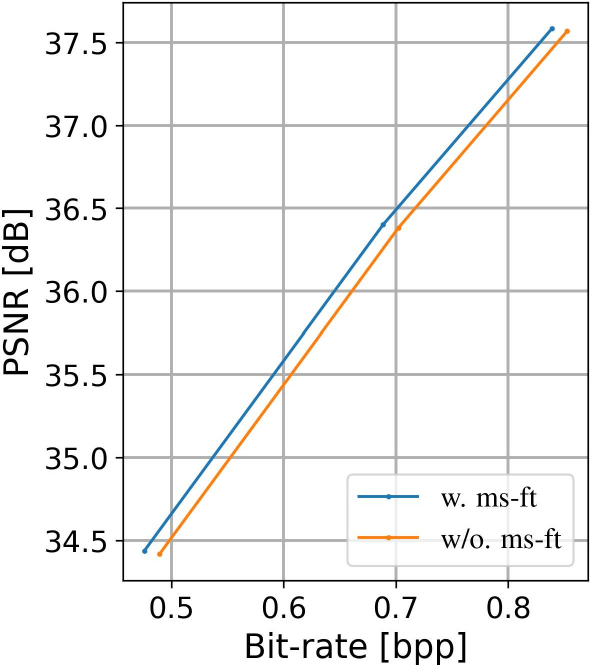}
    \label{subfig:ab_msft}
    }
    \end{subfigure}
\caption{{Comparative study of token-Mixer order, autoregression order, position embedding, and multi-step fine-tuning}. In this table, ``i-c'' vs. ``c-i'' represent inner-group vs. cross-group token-mixer orders. ``sfo'' (spatial-first-order) and ``cfo'' (channel-first-order) refer to different autoregression orders. ``Rel'', ``Abs'' and ``DRel'' is relative, absolute and diamond-shape relative position embedding. ``ms-ft'' stands for multi-step fine-tuning.}
\label{fig:analysis_study}
\end{figure}

\subsubsection{Token-Mixer Order Study\label{subsubsec:block_order_study}}
We design two different token-mixers to construct the GroupedMixer Module. To investigate the most efficient way to combine them, we evaluated the model using 2 permutations. 
As depicted in Figure \ref{subfig:ab_order}, it is beneficial to position the cross-group token-mixer prior to the inner-group token-mixer, as opposed to the converse arrangement.
This preference may be attributed to the fact that tokens in cross-group token-mixer, whether derived  from their proximate spatial positions or otherwise, demonstrate an increased capacity for aggregating more complex information.

\begin{table}[t]
    \centering
    \caption{Performance of different depths and dimensions for transformer blocks used in GroupedMixer. 
    }
    \begin{tabular}{cccccc}
    \toprule
    \textbf{Model} & \textbf{Depth} & \textbf{Dimension} & \textbf{BD-Rate(\%)}  & \textbf{Param.} \\
         \midrule
          \multirow{5}{*}{$5\times2$} 
          &3 & 96 & 11.92 &  12.5M \\
          &3 & 192 & 4.51 &  14.5M\\
          &1 & 384 & 8.98 & 15.4M\\
          &3 & 384 & 2.83 & 22.6M\\
          &6 & 384 & 0.00 & 36.8M\\
         \bottomrule
    \end{tabular}
    \label{tab:depth_heads}
\end{table}

\subsubsection{Spatial-Channel Order\label{subsubsec:sfo_vs_cfo}}
Partitioning the latent variables into groups enables two different autoregression orders, namely,  spatial-first-order (sfo) and channel-first-order (cfo), which can be expressed as follows respectively:
\begin{equation}
    \begin{aligned}
    f_{\text{sfo}}&: \mathbb{R}^{H\times W\times C} \rightarrow \mathbb{R}^{{k_c k_h k_w}\times \frac{H}{k_h} \frac{W}{k_w} \times \frac{C}{k_c}}  \\
    f_{\text{cfo}}&: \mathbb{R}^{H\times W\times C} \rightarrow \mathbb{R}^{{k_h k_w k_c}\times \frac{H}{k_h} \frac{W}{k_w} \times \frac{C}{k_c}}  \\
    \end{aligned}
\end{equation}
Different coding orders prioritize contextual information from different dimensions (spatial or channel), which may result in different compression performance. 
The comparison result is illustrated in Figure~\ref{subfig:ab_order}.
On average, the channel-first-order (cfo) performs almost the same as the model in spatial-first-order (sfo) configuration. We speculate that two different partition methods provide two almost equivalent contextual information in the amortized sense. We adopt the spatial-first-order (sfo) in our final model.

\subsubsection{Position Embedding\label{subsubsec:position_embedding}}
We first examine the effects of position embedding used in cross-group token-mixer. We build models with different position encoding methods, including absolute position embedding, 3D relative position embedding and diamond-shape 3D relative position embedding, which extends diamond position embedding proposed in \cite{qian2022entroformer} from 2D to 3D. The results are presented in Figure~\ref{subfig:ab_pos}.
We observe that the  3D relative position embedding outperforms the others. Moreover, we notice that introducing region constraint in cross-group token-mixer is neither necessary nor effective. Hence, we adopt 3D relative position embedding in our final model. 
In inner-group token-mixer, PEG plays a vital role in generalizing model to high resolution images. We also conduct an experiment to verify this, where configuration follows \ref{subsec:training_details}. 
We compare PEG with previous proposed effective 2D position embedding, diamond position embedding proposed in~\cite{qian2022entroformer}.
As shown in Figure~\ref{subfig:ab_pos_peg},  
even though the two models, one with PEG and the other without PEG, perform very closely on the Kodak dataset, they show significant differences on the larger resolution dataset CLIC'21 Test, with the model without PEG performing much worse than the one with PEG.
This is because PEG is implemented based on depth-wise convolution, which allows for better generalization across images of varying scales compared to diamond position embedding.

\subsubsection{Multi-step Fine-tuning\label{subsubsec:msft}}

To assess the impact of multi-step fine-tuning on training performance, an ablation study on multi-step fine-tuning was conducted, where configuration follows~Section \ref{subsec:training_details}, as shown in Figure~\ref{subfig:ab_msft} In this comparison, ``w. ms-ft'' represents the outcome with the final multi-step fine-tuning, while ``w/o. ms-ft'' corresponds to the results where the third stage was replaced with a one-pass training approach. It was observed that employing multi-step fine-tuning in the training process effectively narrowed the gap between training and testing performance, leading to an overall improvement. Quantitatively, this approach resulted in a reduction of 1.39\% in BD-Rate, underscoring the effectiveness of multi-step fine-tuning in enhancing model performance.
\subsubsection{Depth and Embedding Dimension} 
In Table~\ref{tab:depth_heads} we conduct architecture ablations to examine the relationships between model architecture and compression performance. We explore the following architectural changes: decreasing the number of hidden dimension, holding model depth relatively constant $L=3$; decreasing model depth, holding the hidden dimension constant $d_e=384$. The head dimension $d_h$ is consistently set to a constant $32$.
Our results show that decreasing the hidden dimension size leads to performance deterioration, causing 1.68 \% BD-Rate increase from $d_e=384$ to $d_e=192$ and 9.09\% BD-Rate increase from $d_e=384$ to $d_e=96$. Likewise, reducing the model depth impairs the performance, leading to 8.98\% BD-Rate increase from $L=6$ to $L=1$ and 2.83\% BD-Rate increase from $L=6$ to $L=3$. 
This is because both the hidden dimension and the depth are critical factors for the model, which influence the model's capacity to encode the contextual information.

\section{Conclusion\label{sec:conclusion}}

In this paper, we present an entropy model called GroupedMixer, a group-wise autoregressive transformer for learned image compression. 
To construct a spatial-channel context model that enables global reference and also entails lower complexity, we propose to decompose self-attention into two group-wise token-mixers, namely, inner-group and cross-group token-mixers. We then apply context cache optimization to expedite inference speed, which allows faster coding process and multi-step finetuning. We demonstrate that GroupedMixer achieves SOTA compression performance with fast coding speed.

\bibliographystyle{paper}
\bibliography{Library}
\begin{IEEEbiography}
[{\includegraphics[width=1in,height=1.25in,clip,keepaspectratio]{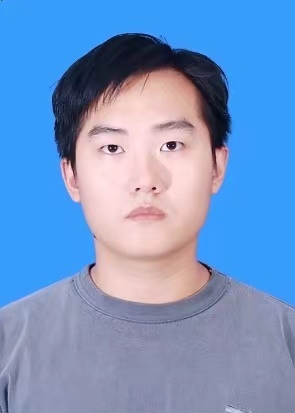}}]{Daxin Li}
received his B.S. degree in the Faculty of Computation from Harbin Institute of Technology, Harbin, China, in 2021. He is currently pursuing his Ph.D. degree in the Faculty of Computation from Harbin Institute of Technology. His research interests lie primarily in the areas of image and video compression, as well as deep learning applications.
\end{IEEEbiography}
\begin{IEEEbiography}[{\includegraphics[width=1in,height=1.25in,clip,keepaspectratio]{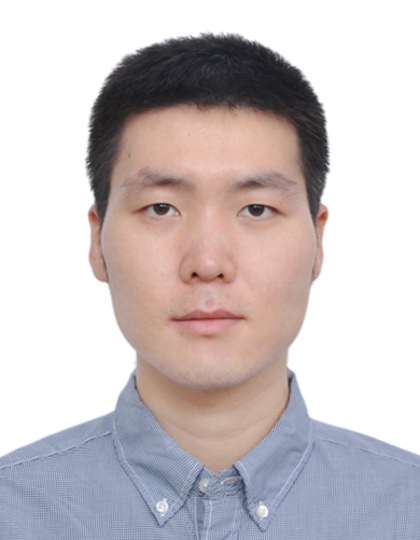}}]{Yuanchao Bai} (Member, IEEE)
received the B.S. degree in software engineering from Dalian University of Technology, Liaoning, China, in 2013. He received the Ph.D. degree in computer science from Peking University, Beijing, China, in 2020. He was a postdoctoral fellow in Peng Cheng Laboratory, Shenzhen, China, from 2020 to 2022. He is currently an assistant professor  with the School of Computer Science and Technology, Harbin Institute of Technology, Harbin, China.
His research interests include image/video compression and processing, deep unsupervised learning, and graph signal processing.
\end{IEEEbiography}
\begin{IEEEbiography}[{\includegraphics[width=1in,height=1.25in,clip,keepaspectratio]{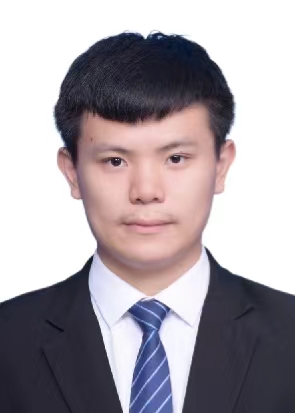}}]{Kai Wang}
received the B.S. degree in software engineering from Harbin Engineering University, Harbin, China, in 2020 and received the M.S. degree of electronic information in software engineering from Harbin Institute of Technology, Harbin, China, in 2022. He is currently pursuing the docter degree in electronic information in Harbin Institute of Technology, Harbin, China.
His research interests include image/video compression and deep learning.
\end{IEEEbiography}
\begin{IEEEbiography}[{\includegraphics[width=1in,height=1.25in,clip,keepaspectratio]{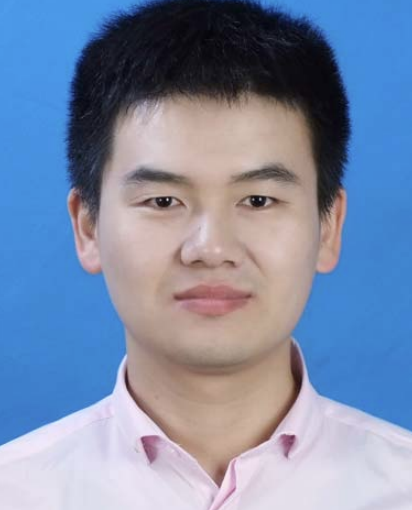}}]{Junjun Jiang} (Senior Member, IEEE) 
received the B.S. degree in mathematics from Huaqiao University, Quanzhou, China, in 2009, and the Ph.D. degree in computer science from Wuhan University, Wuhan, China, in 2014. From 2015 to 2018, he was an Associate Professor with the School of Computer Science, China University of Geosciences, Wuhan. From 2016 to 2018, he was a Project Researcher with the National Institute of Informatics (NII), Tokyo, Japan. He is currently a Professor with the School of Computer Science and Technology, Harbin Institute of Technology, Harbin, China. His research interests include image processing and computer vision.
\end{IEEEbiography}
\begin{IEEEbiography}[{\includegraphics[width=1in,height=1.25in,clip,keepaspectratio]{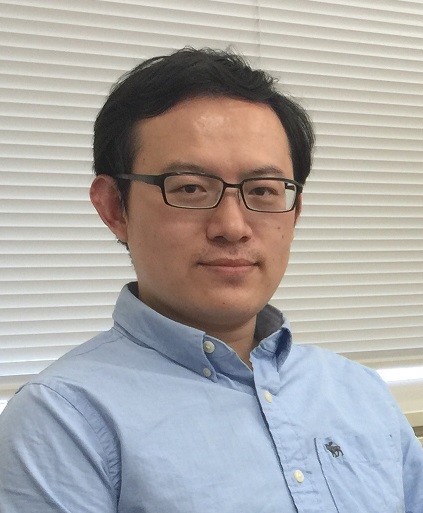}}]{Xianming Liu} (Member, IEEE)
 is a Professor with the School of Computer Science and Technology, Harbin Institute of Technology (HIT), Harbin, China. He received the B.S., M.S., and Ph.D degrees in computer science from HIT, in 2006, 2008 and 2012, respectively. In 2011, he spent half a year at the Department of Electrical and Computer Engineering, McMaster University, Canada, as a visiting student, where he then worked as a post-doctoral fellow from December 2012 to December 2013. He worked as a project researcher at National Institute of Informatics (NII), Tokyo, Japan, from 2014 to 2017. He has published over 60 international conference and journal publications, including top IEEE journals, such as T-IP, T-CSVT, T-IFS, T-MM, T-GRS; and top conferences, such as CVPR, IJCAI and DCC. He is the receipt of IEEE ICME 2016 Best Student Paper Award.
\end{IEEEbiography}
\begin{IEEEbiography}[{\includegraphics[width=1in,height=1.25in,clip,keepaspectratio]{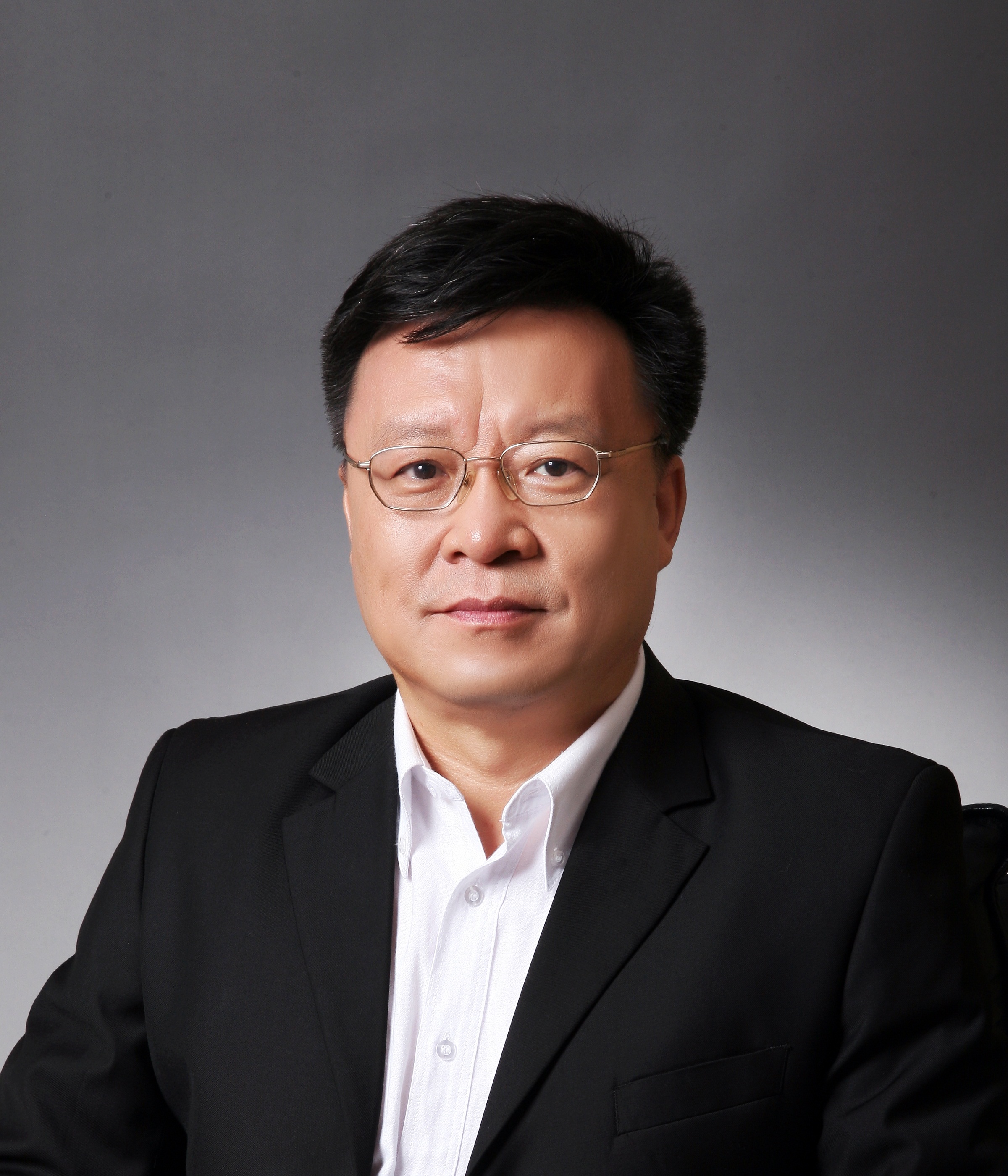}}]{Wen Gao} (Fellow, IEEE) received the Ph.D. degree in electronics engineering from The University of Tokyo, Japan, in 1991. He is currently a Boya Chair Professor in computer science at Peking University. He is the Director of Peng Cheng Laboratory, Shenzhen. Before joining Peking University, he was a Professor with Harbin Institute of Technology from 1991 to 1995. From 1996 to 2006, he was a Professor at the Institute of Computing Technology, Chinese Academy of Sciences. He has authored or coauthored five books and over 1000 technical articles in refereed journals and conference proceedings in the areas of image processing, video coding and communication, computer vision, multimedia retrieval, multimodal interface, and bioinformatics. He served on the editorial boards for several journals, such as ACM CSUR, IEEE TRANSACTIONS ON IMAGE PROCESSING (TIP), IEEE TRANSACTIONS ON CIRCUITS AND SYSTEMS FOR VIDEO TECHNOLOGY (TCSVT), and IEEE TRANSACTIONS ON MULTIMEDIA (TMM). He served on the advisory and technical committees for professional organizations. He was the Vice President of the National Natural Science Foundation (NSFC) of China from 2013 to 2018 and the President of China Computer Federation (CCF) from 2016 to 2020. He is the Deputy Director of China National Standardization Technical Committees. He is an Academician of the Chinese Academy of Engineering and a fellow of ACM. He chaired a number of international conferences, such as IEEE ICME 2007, ACM Multimedia 2009, and IEEE ISCAS 2013.
\end{IEEEbiography}

\end{document}